\documentclass[a4paper,number]{elsarticle}
\pdfoutput=1
\usepackage[utf8]{inputenc}
\usepackage{amssymb,amsmath,array,amsthm}
\usepackage{url}
\usepackage{algorithm,algorithmic}
\biboptions{sort}

\newcommand{\R}[1]{\ensuremath{\mathbb{R}^{#1}}}

\journal{Neurocomputing}

\begin{document}

\begin{frontmatter}
\title{Exploratory Analysis of Functional Data via Clustering and Optimal Segmentation}

\author[enst]{Georges Hébrail}
\ead{Georges.Hebrail@telecom-paristech.fr}
\author[lamsade]{Bernard Hugueney}
\author[inria]{Yves Lechevallier}
\ead{Yves.Lechevallier@inria.fr}
\author[enst]{Fabrice Rossi\corref{cor}}
\ead{Fabrice.Rossi@telecom-paristech.fr}

\address[lamsade]{LAMSADE, Université Paris Dauphine, 
Place du Maréchal de Lattre de Tassigny, 75016 Paris -- France}

\address[enst]{BILab, Télécom ParisTech, LTCI - UMR CNRS 5141 
46, rue Barrault, 75013 Paris -- France}

\address[inria]{Projet AxIS, INRIA, Domaine de Voluceau, Rocquencourt,
  B.P. 105 78153 Le Chesnay Cedex -- France}

\cortext[cor]{Corresponding author}

\begin{abstract}
We propose in this paper an exploratory analysis algorithm for functional
data. The method partitions a set of functions into $K$ clusters and represents
each cluster by a simple prototype (e.g., piecewise constant). The total
number of segments in the prototypes, $P$, is chosen by the user and optimally
distributed among the clusters via two dynamic programming algorithms. The
practical relevance of the method is shown on two real world datasets. 
\end{abstract}

\begin{keyword}
  Functional Data\sep Multiple time series\sep Exploratory analysis\sep
  Clustering \sep  Segmentation \sep Dynamic programming
\end{keyword}

\end{frontmatter}

\section{Introduction}
Functional Data Analysis \cite{RamsaySilverman97} addresses problems in which
the observations are described by functions rather than finite dimensional
vectors. A well known real world example of such data is given by spectrometry
in which each object is analysed via one spectrum, that is a function which
maps wavelengths to absorbance values. Online monitoring of hardware is also a
good example of such data: each object is described by several time series
associated to physical quantities monitored at specified sampling rates.

Another application domain is related to electric power consumption curves,
also called (electric) load curves. Such data describes the electric power
consumption over time of one household or one small or large industry. Load
curves can be very voluminous since there are many consumers (over 35 millions
in France) which can be observed during a long period (often several years)
and at a rate up to one point every minute. Consequently, there is a strong
need for applying unsupervised learning methods to summarize such datasets of
load curves. A typical way to do so is to split each curve into daily (or
weekly) periods and perform a clustering of such daily curves. The motivation
of such analyses is related to several practical problems: understanding the
consumption behaviour of consumers in relation to their equipment or weather
conditions, defining new prices, optimizing the production of electric power,
and in the next years monitoring the household consumption to face high
peaks. 

We focus in this paper on the exploratory analysis of a set of curves (or time
series). The main idea is to provide the analyst with a summary of the set
with a manageable complexity. A classical solution for multivariate data
consists in using a prototype based clustering approach: each cluster is
summarized by its prototype. Standard clustering methods such as K means and
Self Organizing Map have been adapted to functional data and could be used to
implement this solution
\cite{Abraham2000,CottrellGirardRousset1998,DebregeasHebrail1998Curves,RossiConanGuezElGolliESANN2004SOMFunc}. Another
possibility comes from the symbolization approaches in which a time series is
represented by a sequence of symbols. In the SAX approach \cite{DBLP:conf/dmkd/LinKLC03}, the time series is transformed into a piecewise representation with
contiguous time intervals of equal size: the value associated with each
interval is the average of actual values in the interval. This approach is
very fast but does not give any guarantee on the associated error. In
\cite{HugueneyPKDD2006}, a piecewise constant approximation of a time series
is constructed via a segmentation of the time domain into contiguous intervals
on which the series is represented by its average value, which can be turned
into a label in a subsequent quantization step. When we are given a set of
curves, an unique segmentation can be found in order to represent all the
curves on a common piecewise constant basis (see
\cite{RossiLechevallier2008SFC} for an optimal solution). This was used as a
preprocessing step in
e.g. \cite{RossiEtAl06CilsBspline,KrierEtAl2007CILSFDProj}.

We propose in this paper to merge the two approaches: we build a K means like
clustering of a set of functions in which each prototype is given by a simple
function defined in a piecewise way. The input interval of each prototype is
partitioned into sub-intervals (segments) on which the prototype assumes a
simple form (e.g., constant). Using dynamic programming, we obtain an optimal
segmentation for each prototype while the number of segments used in each
cluster is also optimally chosen with respect to a user specified total number
segments. In the case of piecewise constant prototypes, a set of functions is
summarized via $2P-K$ real values, where $K$ is the number of prototypes and
$P$ the total number of segments used to represent the prototypes.

The rest of this paper is organized as follows. Section
\ref{section:one:function} introduces the problem of finding a simple summary
of a single function, links this problem with optimal segmentation and
provides an overview of the dynamic programming solution to optimal
segmentation. Section \ref{section:several:function} shows how single function
summarizing methods can be combined to clustering methods to give a summary of
a set of functions. It also introduces the optimal resource allocation
algorithm that computes an optimal number of segments for each
cluster. Section \ref{secExperiments} illustrates the method on two real world 
datasets.  

\section{Summarizing one function via optimal segmentation: a state-of-the-art}\label{section:one:function}
The proposed solution is built on two elements that are mostly independent:
any standard clustering algorithm that can handle functional data and a
functional summarizing technique that can build an appropriate low complexity
representation of a set of homogeneous functions. The present section
describes the summarizing technique for a single function; the
extension to a set of functions is described in Section
\ref{section:several:function}. 

Building a summary of a function is deeply connected to building an optimal
segmentation of the function, a problem which belongs to the general task of
function approximation. Optimal segmentation has been studied under different
point of views (and different names) in a large and scattered literature (see 
\cite{Stone1961,Bellman1961Function,LechevallierContrainte1976,LechevallierContrainte1990,PicardEtAl2007,AugerLawrence1989,JacksonEtAl2005}
and references therein). The goal of the present section is not to provide new
material but rather to give a detailed exposition of the relations between
summary and segmentation on the one hand, and of the optimal segmentation
framework itself on the other hand. 

\subsection{Simple functions}\label{subsection:simple:function}
We consider a function $s$ sampled in $M$ distinct points $(t_k)_{k=1}^M$ from
the interval $[t_1,t_M]$ (points are assumed to be ordered, i.e.,
$t_1<t_2<\ldots<t_M$). Our goal is to 
approximate $s$ by a \emph{simpler} function $g$ on $[t_1,t_M]$. The
simplicity concept targeted in this article is not based on smoothness as in
regularization \cite{TikhonovArsenin1979} or on capacity as in learning theory
(see e.g., \cite{DevroyeEtAl1996Pattern}). It corresponds rather to an
informal measure of the simplicity of the visual representation of $g$ for a
human analyst.  

The difference between those three aspects (smoothness, capacity and visual
simplicity) can be illustrated by an elementary example: let us assume that
$g$ is chosen as a linear combination of $N$ fixed  functions, i.e.,
\begin{equation}
g(t)=\sum_{i=1}^N\beta_i\phi_i(t).
\end{equation}
It is well known that the set of indicator functions based on functions of the
previous form has a VC dimension of $N$, as long as the functions
$(\phi_i)_{i=1}^N$ are linearly independent (see  e.g.,
\cite{DevroyeEtAl1996Pattern}). If we consider now the $L_2$ norm of the
second derivative of $g$ as a roughness measure, it is quite clear that the
actual smoothness will depend both on the functions $(\phi_i)_{i=1}^N$
and on the values of the coefficients $(\beta_i)_{i=1}^N$. As an
extreme case, one can consider $\phi_1(t)=\text{sign}(t)$ while
the $(\phi_i)_{i=2}^N$ are smooth functions. Then any $g$ with
$\beta_1\neq 0$ is non smooth while all other functions are. 

\begin{figure}[htbp]
  \centering
\includegraphics[width=0.45\textwidth]{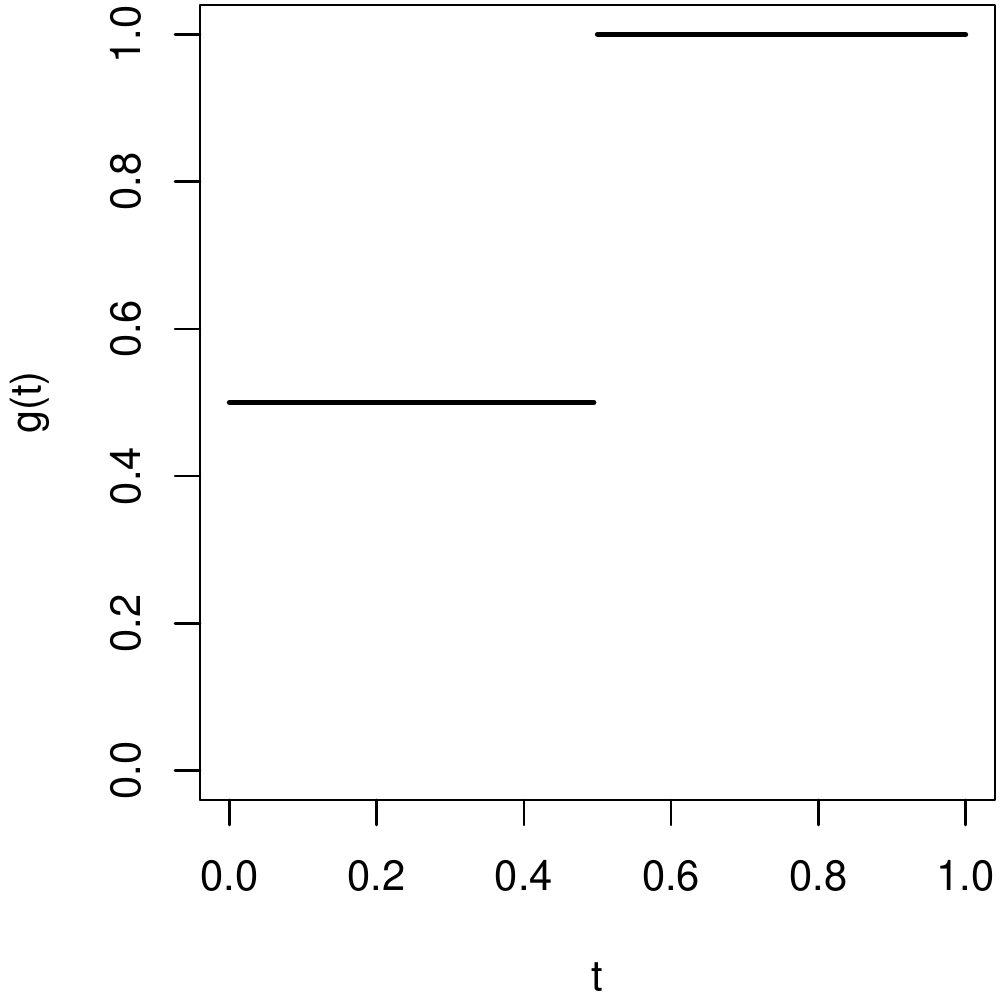}\hfill \includegraphics[width=0.45\textwidth]{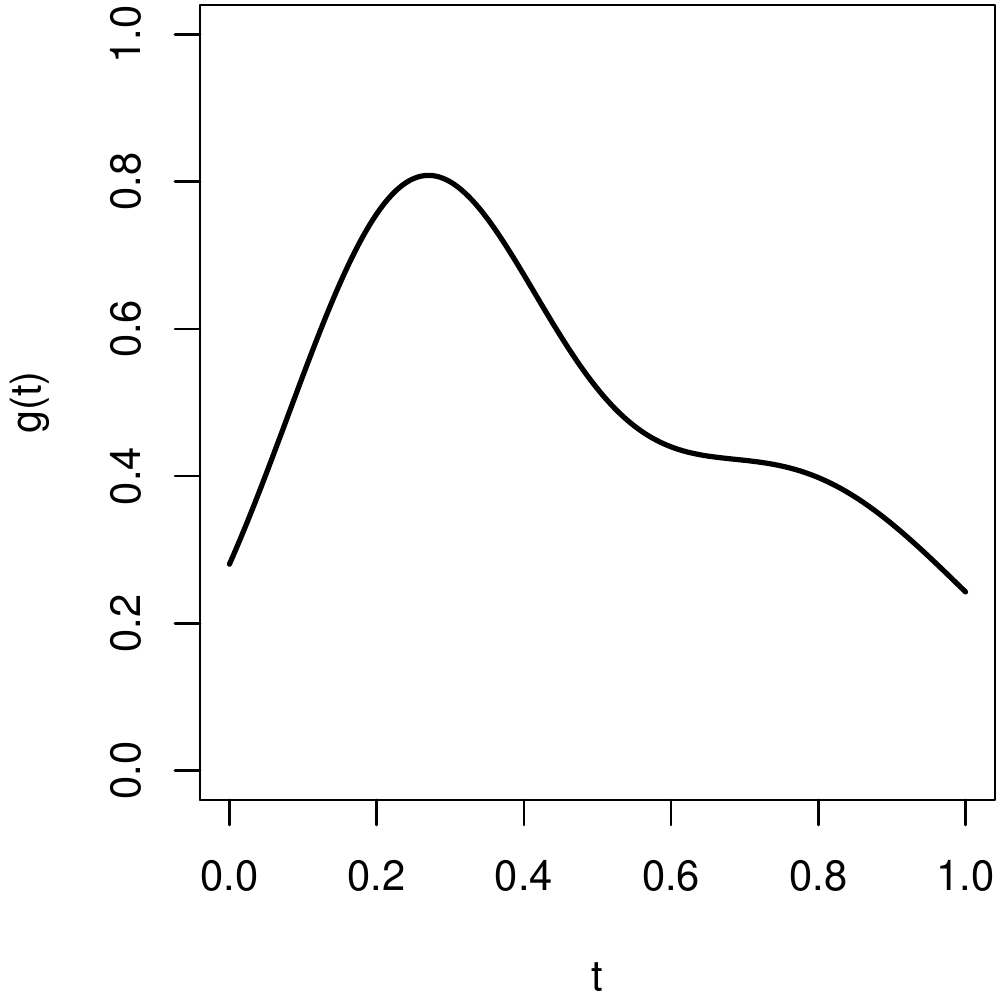}
  \caption{Left: function generated by $(\phi_1,\phi_2)$; right: function generated by $(\psi_1,\psi_2)$}
\label{fig:SmoothnessVsComplexity}
\end{figure}

The visual complexity of $g$ will also clearly depend on both the basis
functions and the coefficients, but in a way that will be generally quite
opposite to the smoothness. Let us again consider an extreme case with
$N=2$, the interval $[0,1]$ and two choices of basis functions,
$(\phi_1,\phi_2)$ and $(\psi_1,\psi_2)$ defined as follows:
\begin{equation}
\phi_1(t)=
\begin{cases}
1 & \text{if } t<\frac{1}{2}\\
0 & \text{if } t\geq \frac{1}{2}
\end{cases}\mspace{36mu}
\phi_2(t)=
\begin{cases}
0 & \text{if } t<\frac{1}{2}\\
1 & \text{if } t\geq \frac{1}{2}
\end{cases}
\end{equation}
and
\begin{equation}
\psi_1(t)=\exp{\left(-16\left(t-\frac{1}{4}\right)^2\right)}\mspace{36mu}\psi_2(t)=\exp{\left(-8\left(\frac{3}{4}-t\right)^2\right)}.
\end{equation}
Obviously, functions generated by $(\phi_1,\phi_2)$ are far less smooth than
functions generated by $(\psi_1,\psi_2)$ (the former are piecewise
constant). However, as shown on Figure \ref{fig:SmoothnessVsComplexity},
functions generated by $(\phi_1,\phi_2)$ are much easier to describe and to
understand than functions generated by $(\psi_1,\psi_2)$.
Indeed, piecewise constant functions admit simple textual description: the
function on the left part of Figure \ref{fig:SmoothnessVsComplexity} takes
values 0.5 for the first half of the period of study and then switches to 1
for the last part. In the contrary, smooth functions are inherently more
complex to describe: the function on the right part of Figure
\ref{fig:SmoothnessVsComplexity} needs a long textual description specifying
its evolution trough time. In addition, no reasonable textual description will
give a way to reconstruct exactly the original function. 

To summarize, classes of functions with identical capacity (as measured by the
VC dimension) can contain functions of arbitrary smoothness and arbitrary
unrelated complexity. We need therefore a specific solution to induce a simple
approximation of a function. Experimental studies in information visualization
(see, e.g., \cite{HealeyEtal1995Preattentive}) have shown that different tasks
(e.g., clustering, change detection, etc.) have different type of adapted
visual features. In our context, we target applications in which a textual
(written or spoken) description of behavior of the function is useful. For
instance in mass spectrometry, peaks positions are very informative and
correspond to descriptions such as ``there is a peak of this height at that
position''. In the case of electrical consumption, the mean level consumption
during a time frame is an interesting aspect and corresponds to description
such as ``the outlet consumes this amount of power from time $a$ to time
$b$''. Hardware monitoring provides another example in which quantities of
interest are generally constant on some time frames and switch linearly from
one stable value to another. 

To implement this idea, we rely on a segmentation approach detailed in the
next section (and covered in a large body of literature, as recalled in the
introduction). A \emph{simple} function in our context is given by a partition
of $[t_1,t_M]$ into a small number of segments on which the function takes a
simple application dependent parametric form. If the parametric form is
reduced to a single value, the simple function is piecewise constant and is
easy to describe in text. Other convenient parametric forms are affine
functions (to provide less jumpy transitions) and peaks. This approach is
closely related to (and inspired by) symbolization techniques used for
instance in time series indexing \cite{HugueneyPKDD2006}: the idea is to build
a piecewise constant approximation of a function and then to quantize the
levels (the constants) into a small number of values. The obtained values are
replaced by symbols (e.g., letters) and the time series is therefore
translated into a sequence of symbols.

\subsection{Segmentation}\label{subsection:segmentation}
Let us analyze first an example of the general scheme described in the
previous section. We consider a piecewise constant approximation given by $P$
intervals $(I_p)_{p=1}^P$ and the corresponding $P$ values $(a_p)_{p=1}^P$ (we
assume that $(I_p)_{p=1}^P$ is a partition of $[t_1,t_M]$). The approximating
function $g$ is defined by $g(t)=a_p$ when $t\in I_p$. If we consider the
$L_2$ error criterion, we obtain a particular case of the Stone approximation
problem \cite{Stone1961} as we have to minimize
\begin{equation}
  \label{eq:optimal:segmentation:one}
\mathcal{E}\left(s,(I_p)_{p=1}^P,(a_p)_{p=1}^P\right)=\int_{t_1}^{t_M}(s(t)-g(t))^2dt=\sum_{p}\int_{I_p}(s(t)-a_p)^2dt.
\end{equation}
The integral can be approximated by a quadrature scheme. We consider the
simplest setting in which a summation over $(t_k)_{k=1}^M$ provides a
sufficient approximation, but the generalization to weighted schemes such as
\cite{LeeVerleysenLpWSOM2005} is
straightforward. In practice, we optimize therefore
\begin{equation}
  \label{eq:optimal:segmentation:one:sum}
E\left(s,(I_p)_{p=1}^P,(a_p)_{p=1}^P\right)=\sum_{p}\sum_{t_k\in I_p}(s(t_k)-a_p)^2.
\end{equation}
The difficulty lies in the segmentation, i.e., in the choice of the partition
$(I_p)_{p=1}^P$, as, given this partition, the optimal values of the
$(a_p)_{p=1}^P$ are equal to the $\mu_{p}=\frac{1}{|I_p|}\sum_{t_k\in
  I_p}s(t_k)$, where $|I_p|$ denotes the number of $t_k$ in $I_p$.

More generally, the proposed method is built upon a parametric approximation
on each segment. Let us denote $Q(s,I)$ the minimal error made by this
parametric approximation of $s$ on interval $I$. The piecewise constant
approximation described above corresponds to 
\begin{equation}
  \label{eq:Q:constant}
Q(s,I)=\sum_{t\in I}\left(s(t)-\frac{1}{|I|}\sum_{v\in  I}s(v)\right)^2. 
\end{equation}
Similar expressions can be derived for a piecewise linear approximation or any
other reasonable solution. For instance
\begin{equation}
  \label{eq:Q:constant:L1}
Q(s,I)=\sum_{t\in I}\left|s(t)-\text{m}(s,I)\right|,
\end{equation}
where $\text{m}(s,I)$ is the median of $\{s(t_k)|t_k\in I\}$, provides a more
robust solution. 

Given $Q$, the optimal segmentation of $s$ is obtained by minimizing
\begin{equation}
  \label{eq:optimal:segmentation:one:sum:generic}
E\left(s,(I_p)_{p=1}^P\right)=\sum_{p}Q(s,I_p),
\end{equation}
over the partitions $(I_p)_{p=1}^P$ of $[t_1,t_M]$. This formulation
emphasizes the central hypothesis of the method: the optimal
approximation on a segment depends only on the values taken by $s$ in the
segment. While a more general segmentation problem, in which this constraint
is removed, can be easily defined, its resolution is much more computationally
intensive than the one targeted here. 

\subsection{Optimal segmentation}\label{secOptimalSegmentation}
Bellman showed in \cite{Bellman1961Function} that the Stone approximation
problem \cite{Stone1961} problem can be solved exactly and efficiently by
dynamic programming: this is a consequence of the independence between the
approximation used in each segment. This has been rediscovered and/or
generalized in numerous occasions (see e.g.,
\cite{PicardEtAl2007,AugerLawrence1989,JacksonEtAl2005}). 

A different point of view is given in
\cite{LechevallierContrainte1976,LechevallierContrainte1990}. As the function
$s$ is known only through its values at $(t_k)_{k=1}^M$, defining a partition
of $[t_1,t_M]$ into $P$ intervals $(I_p)_{p=1}^P$ is not needed. A partition
$(C_p)_{p=1}^P$ of $t_1<t_2<\ldots<t_M$ is sufficient, as long as it is
\emph{ordered}: if $t_k\in C_p$ and $t_l\in C_p$ , then
$\{t_k,t_{k+1},\ldots,t_l\}\subset C_p$. The goal of the segmentation is then
to find an optimal ordered partition of $(t_k)_{k=1}^M$ according to the error
criterion defined in equation \eqref{eq:optimal:segmentation:one:sum} (where
$I_p$ is replaced by $C_p$ and $Q$ is modified accordingly). As shown in
\cite{LechevallierContrainte1976,LechevallierContrainte1990}, this problem can
be solved by dynamic programming as long as the error criterion is
\emph{additive}, i.e., of the general form
\begin{equation}
  \label{eq:AdditiveCriterion}
E\left(s,(C_p)_{p=1}^P\right)=\sum_{p}Q(s,C_p).
\end{equation}

Moreover, the summation operator can be replaced by any commutative aggregation
operator, e.g., the maximum. As already mentioned in the previous section, the
crucial point is that the error criterion is a combination of values obtained
independently on each cluster.

The dynamic programming algorithm for minimizing the cost of equation
\eqref{eq:AdditiveCriterion} is obtained as follows. We define first a set of
clustering problems:
\begin{equation}
F(s,k,j)=\min_{(C_p)_{p=1}^j,\text{ ordered partition of
  }\{t_k,t_{k+1},\ldots,t_M\}}\sum_{p}Q(s,C_p).
\end{equation}
The basic idea is to compute $F(s,k,j)$ using $F(s,.,j-1)$, as the best
partition into $j$ clusters is obtained with
\begin{equation}
  \label{eq:recursion:dp}
F(s,k,j)=\min_{k\leq l\leq M-j+1}Q\left(s,\{t_k,\ldots,t_{l}\}\right)+F(s,l+1,j-1).
\end{equation}
Indeed the partition is ordered and therefore, up to a renumbering of the
clusters, there is $l$ such that $C_1=\{t_1,\ldots,t_{l}\}$. Moreover, the
quality measure is additive and therefore finding the best partition in $j$
clusters with the constraint that $C_1=\{t_1,\ldots,t_{l}\}$ corresponds to
finding the best partition of $\{t_{l+1},\ldots,t_M\}$ in $j-1$ clusters. This
leads to Algorithm \ref{algo:dp:single}, in which $W(k,j)$ is the winner
split, i.e., the $l$ that realizes the minimum in equation
\eqref{eq:recursion:dp}. 
\begin{algorithm}[htbp]
\caption{Dynamic programming summarizing of a single function}
\label{algo:dp:single}
  \begin{algorithmic}[1]
    \FOR{$k=1$ to $M$}
      \STATE $F(s,k,1) \leftarrow Q\left(s,\{t_k,t_{k+1},\ldots,t_M\}\right)$
      \STATE $W(k,1) \leftarrow \text{NA}$ \COMMENT{no meaningful value at for $j=1$}
    \ENDFOR
    \FOR{$j=2$ to $P$}
      \FOR{$k=1$ to $M-j+1$}
         \STATE $F(s,k,j)\leftarrow Q(s,\{t_k\})+F(s,k+1,j-1)$
         \STATE  $W(k,1) \leftarrow k$
         \FOR{$l=k+1$ to $M-j+1$}
            \IF{$Q(s,\{t_k,\ldots,t_l\})+F(s,l+1,j-1)<F(s,k,j)$}
               \STATE $F(s,k,j)\leftarrow Q(\{s,t_k,\ldots,t_l\})+F(s,l+1,j-1)$
               \STATE  $W(k,1) \leftarrow l$
            \ENDIF
         \ENDFOR
      \ENDFOR
    \ENDFOR
    \STATE $C\leftarrow (\text{NA},\ldots,\text{NA})$
    \STATE $C(P-1)\leftarrow W(1,P)$
    \FOR{$j=P-2$ to $2$}
       \STATE $C(j)\leftarrow W(C(j+1)+1,j+1)$
    \ENDFOR
  \end{algorithmic}
\end{algorithm}
The final loop is the backtracking phase in which the split positions are
assembled to produce the result of the algorithm, $C$, which gives the last
indexes of the $P-1$ first clusters of the partition. It should be noted that
while Algorithm \ref{algo:dp:single} outputs only the optimal partition in $P$
clusters, an actual implementation will of course provide in addition the
optimal model. Moreover, as the algorithm produces all values of $F(.,.)$, it
can provide at once all optimal partitions in $p$ clusters for $p$ ranging
from $1$ to $P$. This is done via $P$ backtracking loops for a total cost in
$O(P^2)$. 

\begin{figure}[htbp]
  \centering
\includegraphics[width=0.47\textwidth]{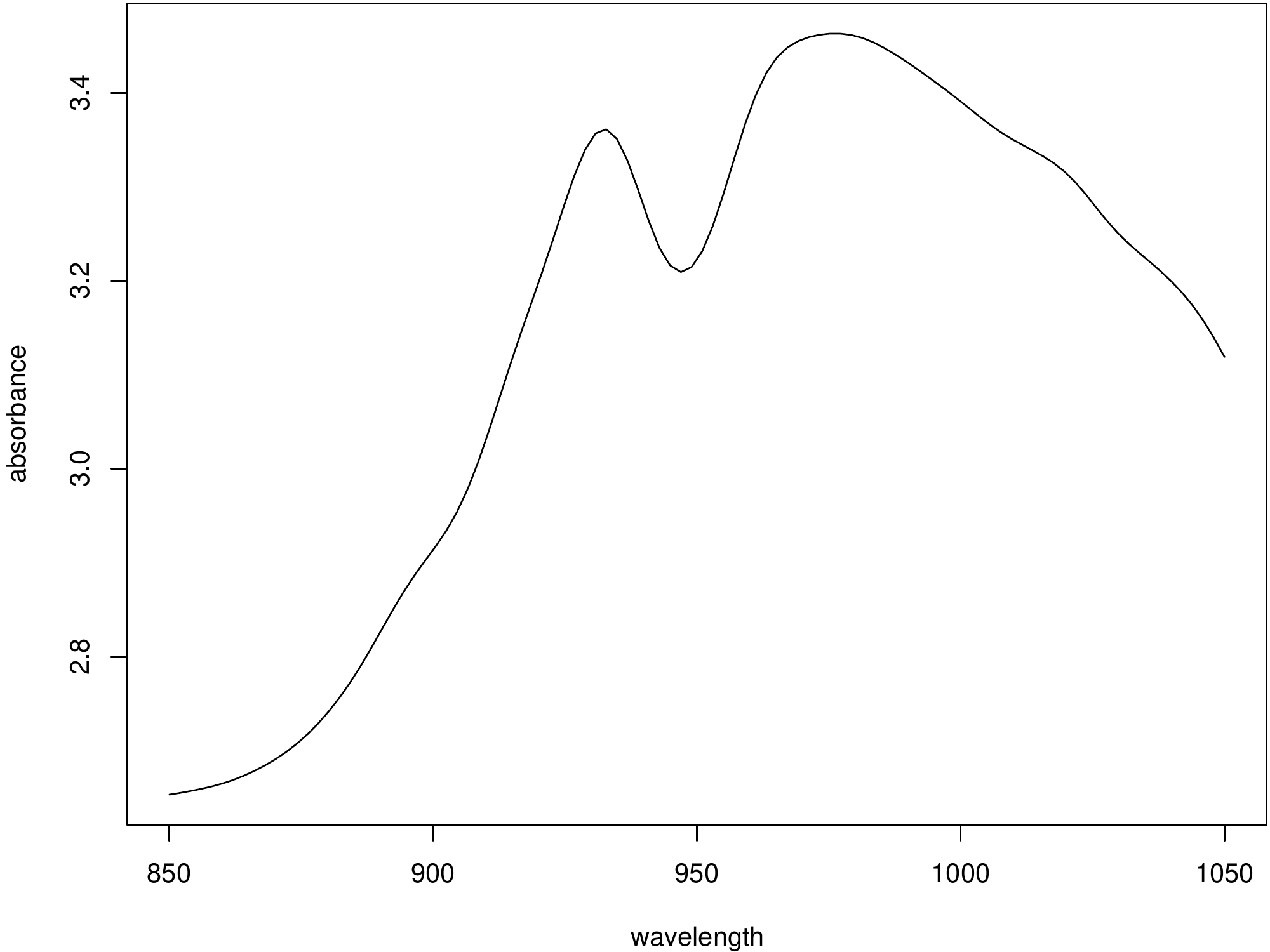} \hfill \includegraphics[width=0.47\textwidth]{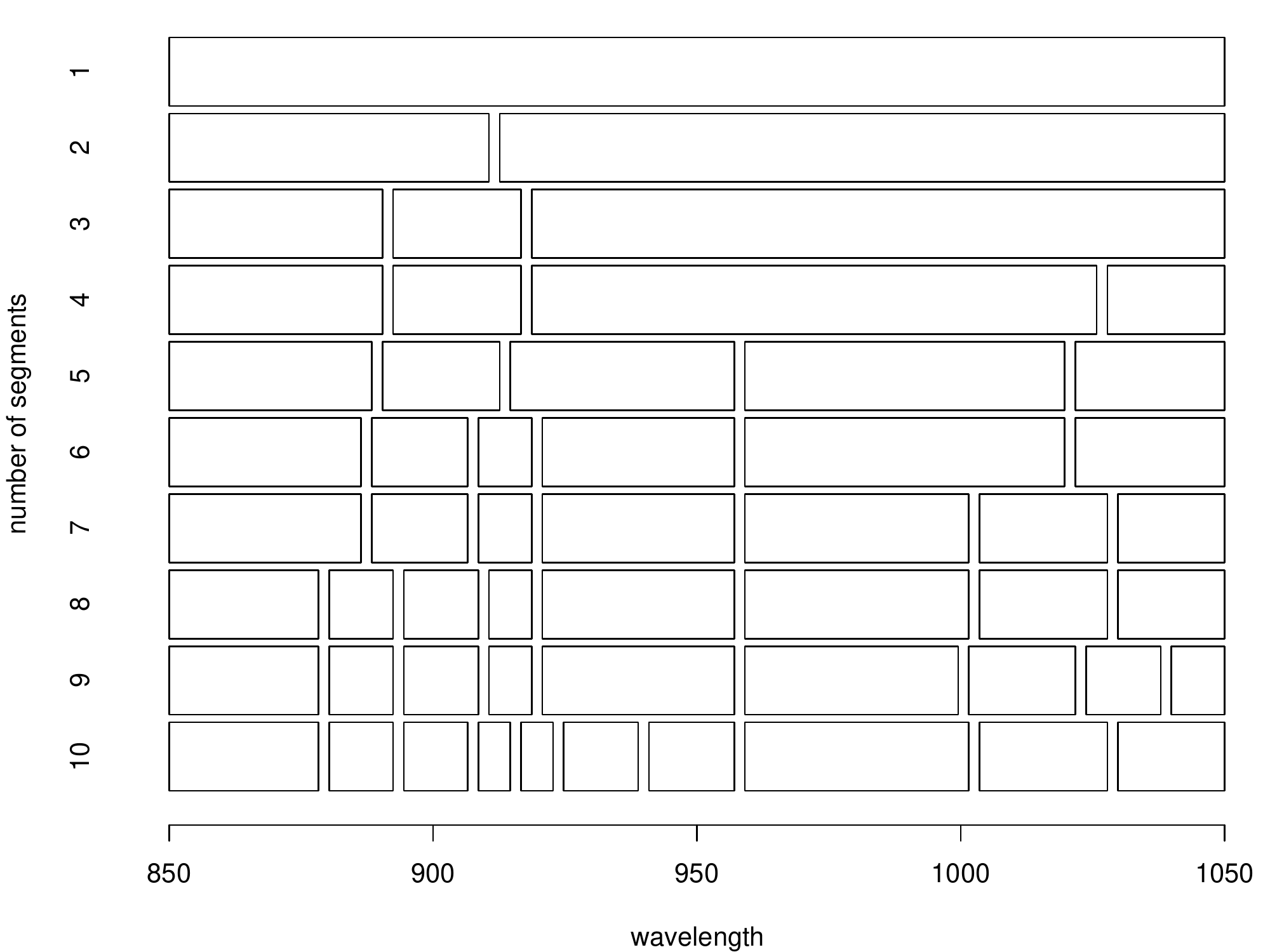}  
  \caption{Segmentation of a single function: original function on the left,
    segment positions on the right}
  \label{fig:Tecator:One}
\end{figure}

\begin{figure}[htbp]
  \centering
\includegraphics[width=0.47\textwidth]{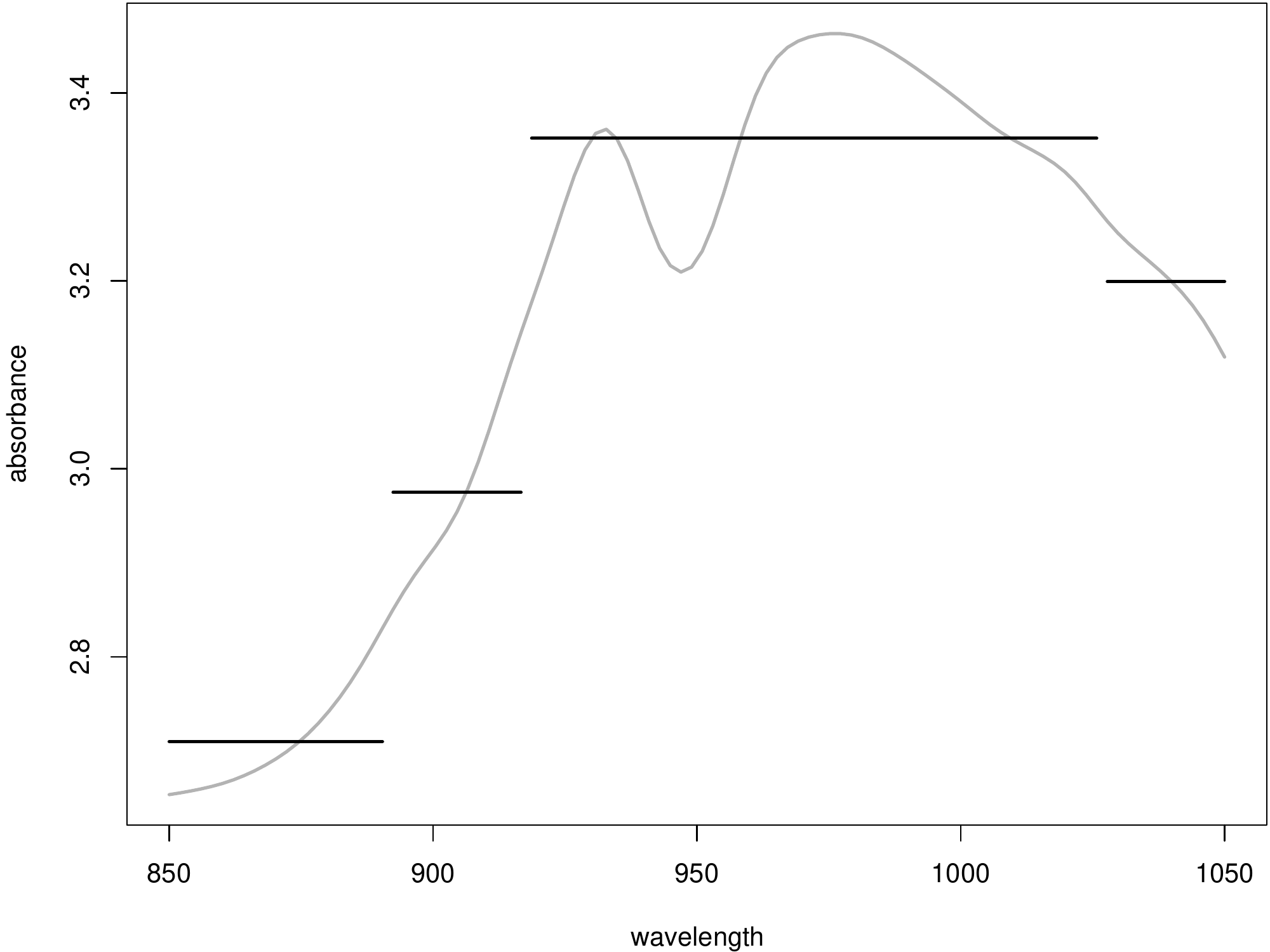} \hfill \includegraphics[width=0.47\textwidth]{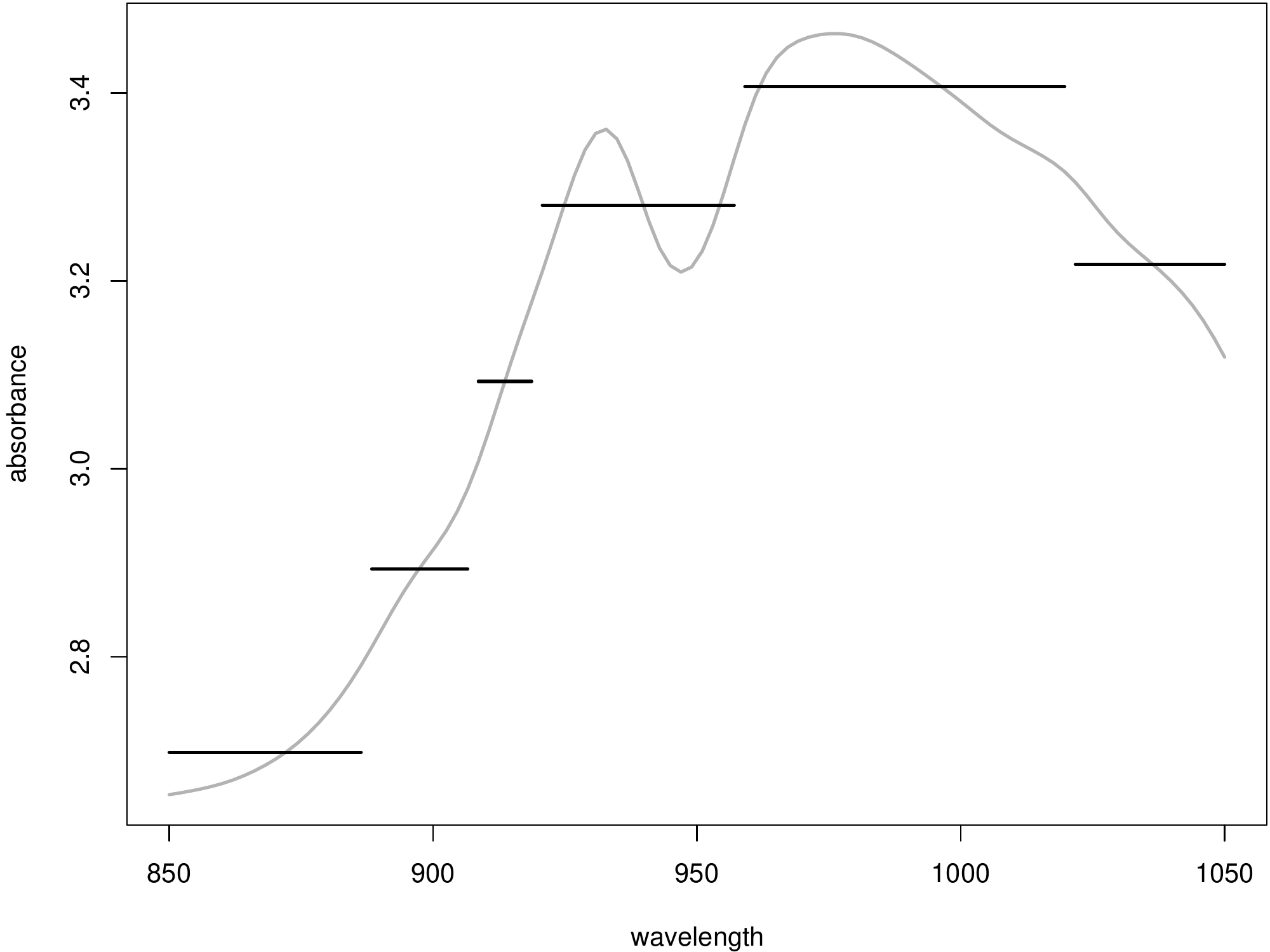}  
  \caption{Piecewise constant approximations with 4 segments on the left and 6
  segments on the right}
  \label{fig:Tecator:Two}
\end{figure}

Figures \ref{fig:Tecator:One} and \ref{fig:Tecator:Two} show an example of the
results obtained by the algorithm. The original function on the left side of
Figure \ref{fig:Tecator:One} is a spectrum from the Tecator
dataset\footnote{Data are available on statlib at
  \url{http://lib.stat.cmu.edu/datasets/tecator} and consist in near infrared
  absorbance spectrum of meat samples recorded on a Tecator Infratec Food and
  Feed Analyzer.} for which $M=100$. The spectrum is segmented into 1 to 10
segments (positions of the optimal segments are given on the right side of
Figure \ref{fig:Tecator:One}). The resulting approximations for 4 and 6
segments are given on Figure \ref{fig:Tecator:Two}.

\subsection{Algorithmic cost}
Given all the $Q\left(s,\{t_{k},\ldots,t_{l}\}\right)$, Algorithm
\ref{algo:dp:single} runs in $O(PM^2)$. This is far more efficient that a
naive approach in which all possible ordered partitions would be evaluated
(there are $\frac{M!}{(M-P)!}$ such partitions). However, an efficient
implementation is linked to a fast evaluation of the
$Q\left(s,\{t_{k},\ldots,t_{l}\}\right)$. For instance, a naive implementation
based on equation \eqref{eq:Q:constant} leads to a $O(M^3)$ cost that
dominates the cost of Algorithm \ref{algo:dp:single}. 

Fortunately, recursive formulation can be used to reduce in some cases the
computation cost associated to the $Q\left(s,\{t_{k},\ldots,t_{l}\}\right)$ to
$O(M^2)$. As an illustration, Algorithm \ref{algo:recursive:constant} computes
$Q$ for equation \eqref{eq:Q:constant} (piecewise constant approximation) in
$O(M^2)$. 
\begin{algorithm}
\caption{Recursive calculation of $Q\left(s,\{t_{k},\ldots,t_{l}\}\right)$}
\label{algo:recursive:constant}
  \begin{algorithmic}[1]
\FOR{$k=1$ to $M$}
  \STATE $\mu(s,\{t_{k}\})\leftarrow s(t_{k})$
  \STATE $Q(s,\{t_{k}\})\leftarrow 0$ 
\ENDFOR
\FOR{$l=2$ to $M$}
  \STATE $\mu(s,\{t_{1},\ldots,t_{l}\})\leftarrow
\frac{1}{l}\left[(l-1)\mu(s,\{t_{1},\ldots,t_{l-1}\})+s(t_l)\right]$
  \STATE $Q(s,\{t_{1},\ldots,t_{l}\})\leftarrow Q(s,\{t_{1},\ldots,t_{l-1}\})
+\frac{l}{l-1}\left[s(t_l)-\mu(s,\{t_{1},\ldots,t_{l}\})\right]^2$
\ENDFOR
\FOR{$k=2$ to $M-1$}
  \FOR{$l=k+1$ to $M$}
    \STATE $\mu(s,\{t_{k},\ldots,t_{l}\})\leftarrow
\frac{1}{l-k+1}\left[(l-k+2)\mu(s,\{t_{k-1},\ldots,t_{l}\})-s(t_{k-1})\right]$
\STATE $Q(s,\{t_{k},\ldots,t_{l}\})\leftarrow Q(s,\{t_{k-1},\ldots,t_{l}\})-\frac{l-k+1}{l-k+2}[s(t_{k-1})-\mu(s,\{t_{k},\ldots,t_{l}\})]^2$
  \ENDFOR
\ENDFOR
  \end{algorithmic}
\end{algorithm}
Similar algorithms can be derived for other choices of the approximation
solution used in each segment. The memory usage can also be reduced from
$O(M^2)$ in Algorithms \ref{algo:dp:single} and \ref{algo:recursive:constant}
to $O(M)$ (see, e.g., \cite{LechevallierContrainte1976,LechevallierContrainte1990}).  

\subsection{Extensions and variations}\label{subsection:extensions}
As mentioned before, the general framework summarized in equation
\eqref{eq:AdditiveCriterion} is very flexible and accommodates numerous
variations. Those variations include the approximation model (constant,
linear, peak, etc.), the quality criterion (quadratic, absolute value, robust
Huber loss \cite{HuberLoss1964}, etc.) and the aggregation operator (sum or
maximum of the point wise 
errors). 

\begin{figure}[htbp]
  \centering
\includegraphics[width=0.47\textwidth]{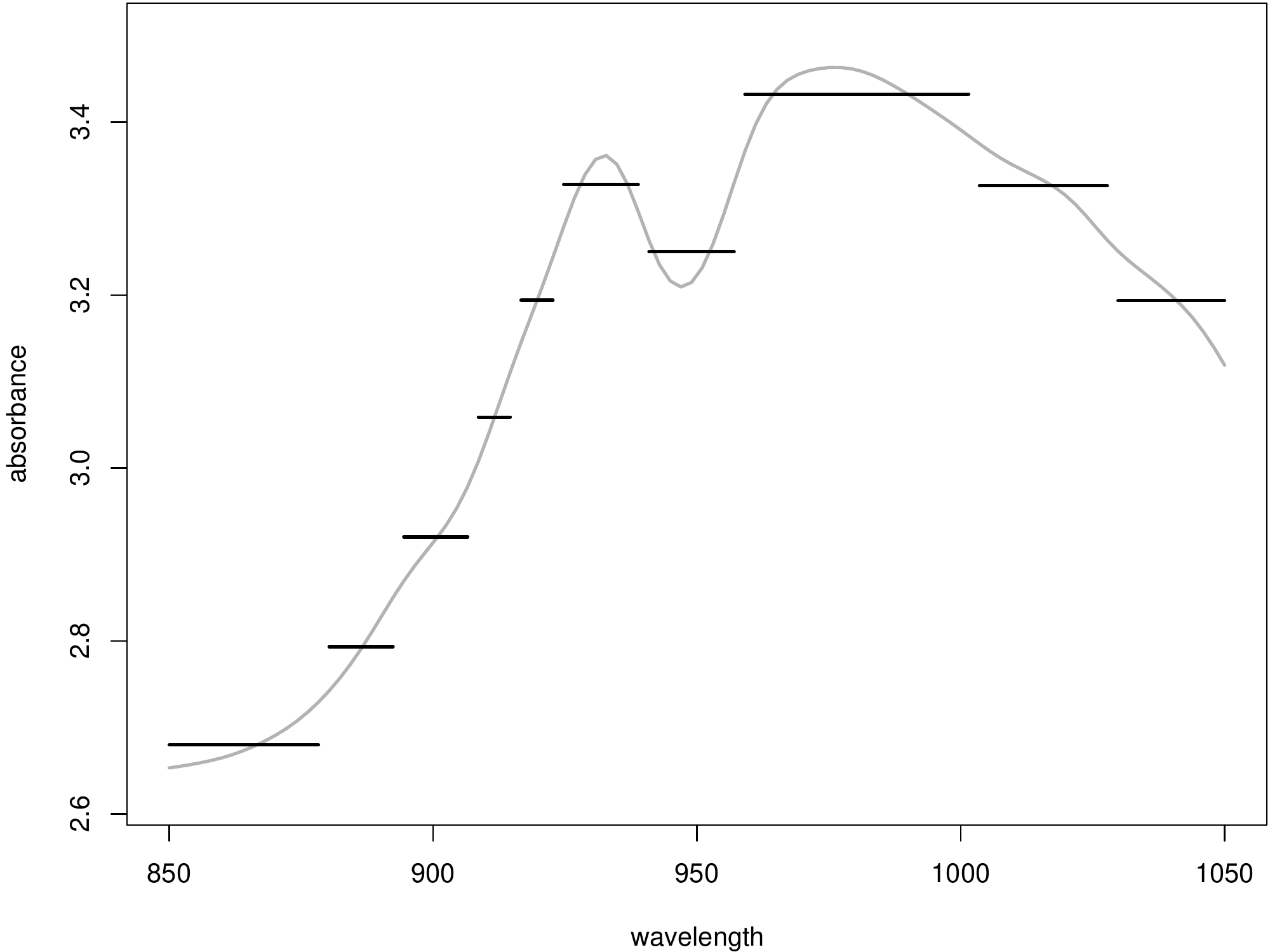} \hfill \includegraphics[width=0.47\textwidth]{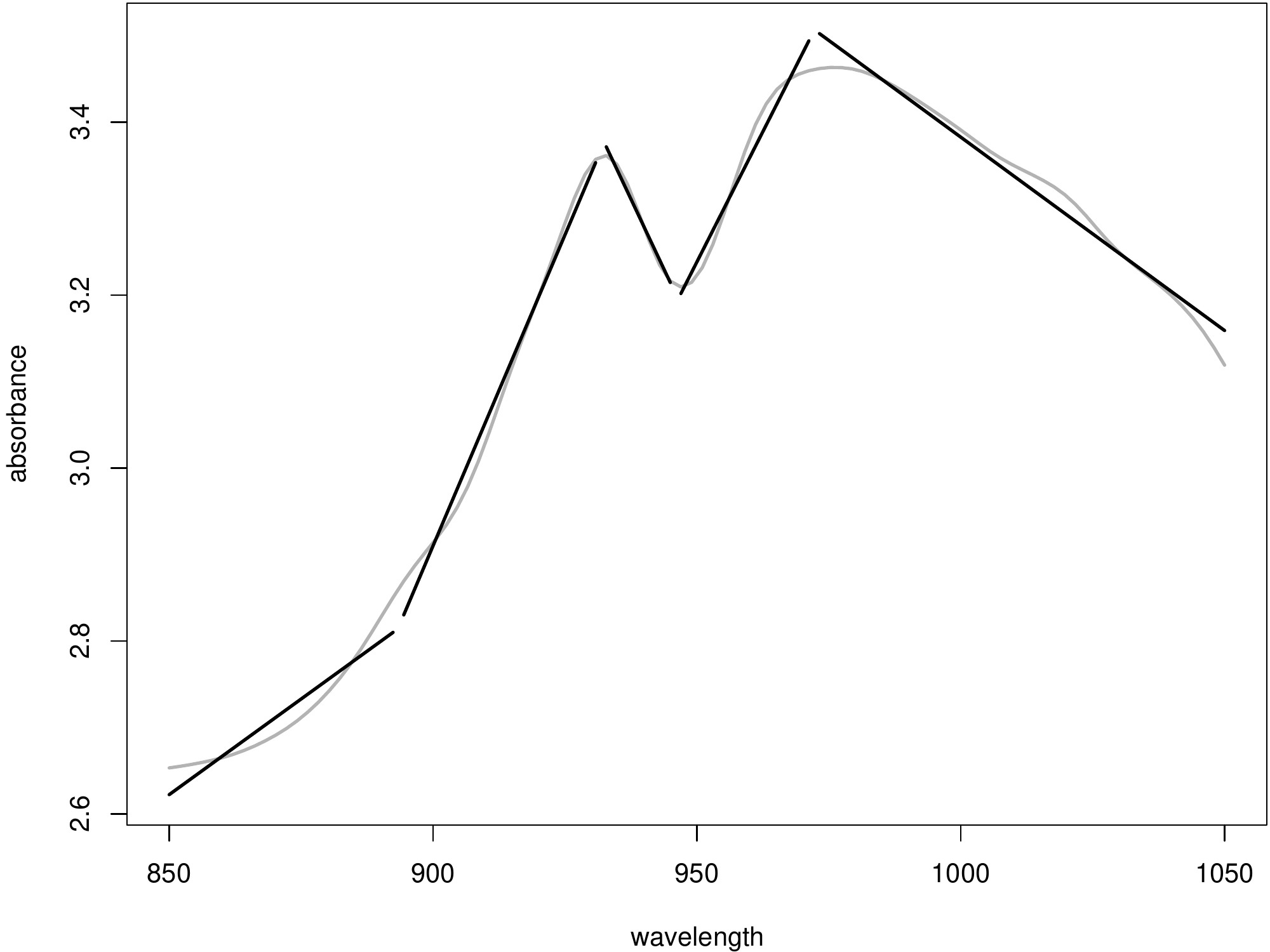}  
  \caption{Segmentation of a single function: piecewise constant with 10
    segments (left) versus piecewise linear with 5 segments (right)}
  \label{fig:Tecator:LinVSConst}
\end{figure}

Figure \ref{fig:Tecator:LinVSConst} gives an application example:
the spectrum from Figure \ref{fig:Tecator:One} is approximated via a piecewise
linear representation (on the right hand side of the Figure): as expected, the
piecewise linear approximation is more accurate than the piecewise constant
one, while the former uses less numerical parameters than the latter. 
Indeed, the piecewise linear approximation is easier to describe than the
piecewise constant approximation: one needs only to specify the 4 breakpoints
between the segments together with 10 extremal values, while the piecewise
constant approximation needs 10 numerical values together with 9 breakpoints.

However, this example is particularly favorable for the piecewise linear
approximation. Indeed, the independence hypothesis embedded in equation
\eqref{eq:AdditiveCriterion} and needed for the dynamic programming approach
prevents us from introducing non local constraints in the approximating
function. In particular, one cannot request for the piecewise linear
approximation to be continuous. This is emphasized on Figure
\ref{fig:Tecator:LinVSConst} by the disconnected drawing of the approximation
function: in fact, the function $g$ is not defined between $t_k$ and $t_{k+1}$
if those two points belong to distinct clusters (in the case of Bellman's
solution \cite{Bellman1961Function}, the function is not well defined at the
boundaries of the intervals). Of course, one can use linear interpolation to
connect $g(t_k)$ with $g(t_{k+1})$, but this can introduce in practice $P-1$
additional very short segments, leading to an overly complex model.

In the example illustrated on Figure \ref{fig:Tecator:LinVSConst}, the
piecewise approximation does not suffer from large jumps between
segments. Would that be the case, one could use an enriched description of the
function specifying the starting and ending values on each segment. This would
still be simpler than the piecewise constant description. However, if the
original function is continuous, using a non continuous approximation can lead
to false conclusion. 

To illustrate the flexibility of the proposed framework beyond the simple
variations listed above, we derive a solution to this continuity problem
\cite{LechevallierContrainte1990}. Let us first define
\begin{equation}
  \label{eq:Q:pseudo:lin}
Q(s,\{t_{k},\ldots,t_{l}\})=\sum_{j=1}^l\left(s(t_j)-\frac{(s(t_k)-s(t_l))t+(t_ks(t_l)-t_ls(t_k))}{t_k-t_l}\right)^2.
\end{equation}
This corresponds to the total quadratic error made by the linear interpolation
of $s$ on $\{t_{k},\ldots,t_{l}\}$ based on the two interpolation points
$(t_k,s(t_k))$ and $(t_l,s(t_l))$. Then we replace the quality criterion of
equation \eqref{eq:AdditiveCriterion} by the following one:
\begin{equation}
  \label{eq:CriterionInterp}
E((k_l)_{l=0}^P)=\sum_{l=1}^PQ(s,\{t_{k_{l-1}},\ldots,t_{k_{l}}\}),
\end{equation}
defined for an increasing series $(k_l)_{l=0}^P$ with $k_0=1$ and
$k_P=M$. Obviously, $E((k_l)_{l=0}^P)$ is the total quadratic error made by
the piecewise linear interpolation of $s$ on $[t_1,t_M]$ based on the $P+1$
interpolation points $(t_{k_l},s(t_{k_l}))_{l=0}^P$. While $(k_l)_{l=0}^P$
does not correspond strictly to a partition of $\{t_1,\ldots,t_{M}\}$ (because
each interpolation point belongs to two segments, expect for the first and the
last ones), the criterion $E$ is additive and has to be optimized on an
ordered structure. As such, it can be minimized by a slightly modified version
of Algorithm \ref{algo:dp:single}. 
\begin{figure}[htbp]
  \centering
\includegraphics[width=0.47\textwidth]{tecator-high-fat-segments-5-lin.pdf} \hfill \includegraphics[width=0.47\textwidth]{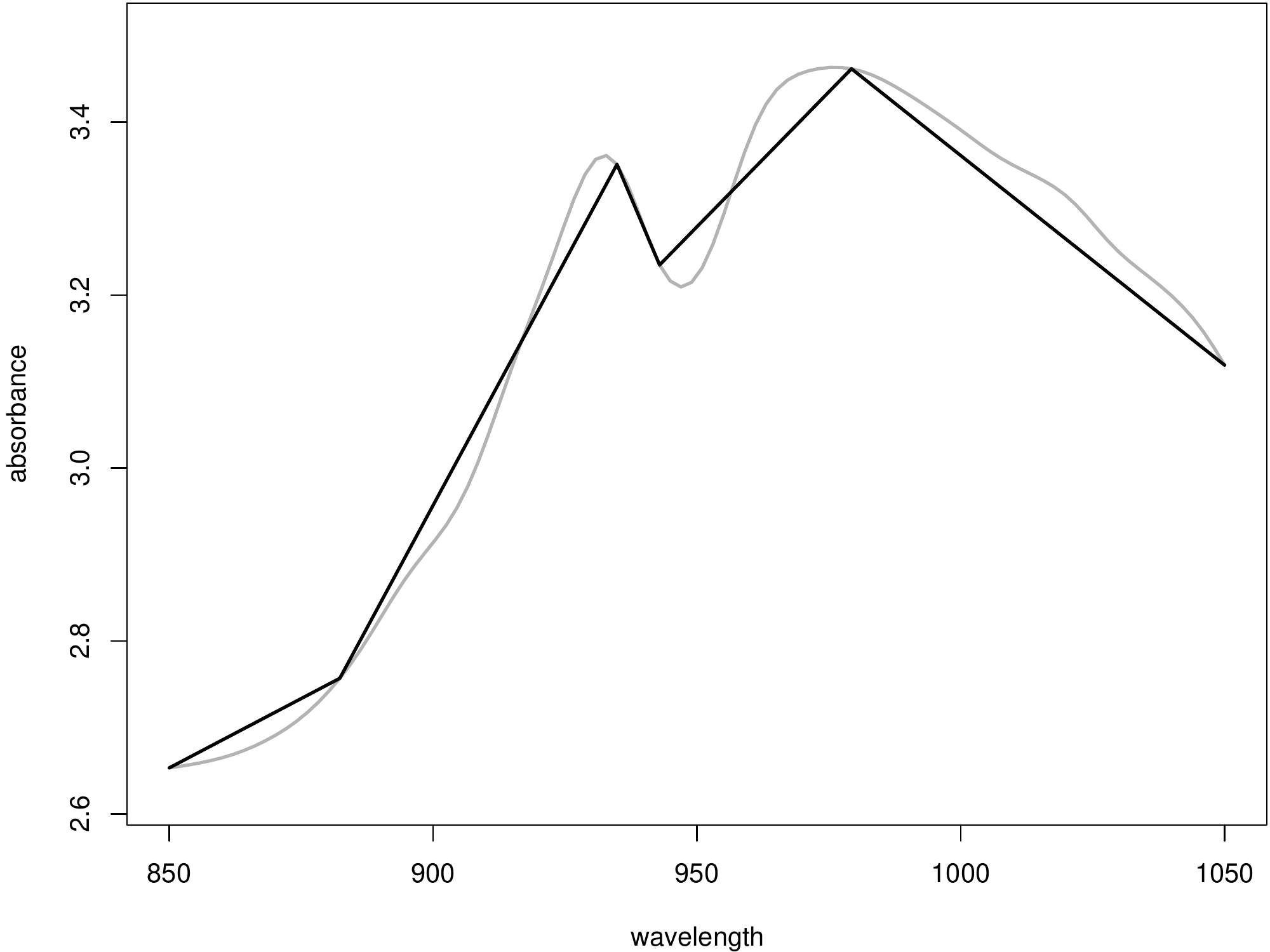}  
  \caption{Segmentation of a single function: piecewise linear with 5 segments
    (left) versus linear interpolation with 5 segments (right)}
  \label{fig:Tecator:LinVSInterp}
\end{figure}
Figure \ref{fig:Tecator:LinVSInterp} shows the differences between the
piecewise linear approach and the linear interpolation approach. For
continuous functions, the latter seems to be more adequate than the former: it
does not introduce arbitrary discontinuity in the approximating function and
provides a simple summary of the original function. 

The solutions explored in this section target specifically the summarizing
issue by providing simple approximating function. As explained in Section
\ref{subsection:simple:function}, this rules out smooth approximation. In
other applications context not addressed here, one can trade simplicity for
smoothness. In \cite{ChamroukhiEtAlESANN2009}, a model based approach is used
to build a piecewise polynomial approximation of a function. The algorithm
provides a segmentation useful for instance to identify underlying regime
switching as well as a smooth approximation of the original noisy
signal. While the method and framework are related, they oppose on the favored
quality: one favors smoothness, the other favors simplicity. 

\section{Summarizing several functions}\label{section:several:function}
Let us now consider $N$ functions $(s_i)_{i=1}^N$ sampled in $M$ distinct
points $(t_k)_{k=1}^M$ from the interval $[t_1,t_M]$ (as in Section
\ref{section:one:function}, the points are assumed to be ordered). Our goal is
to leverage the function summarizing framework presented in Section
\ref{section:one:function} to build a summary of the whole set of
functions. We first address the case of a homogeneous set of functions and
tackle after the general case. 

\subsection{Single summary}\label{subsection:single:summary}
Homogeneity is assumed with respect to the chosen functional metric, i.e., to
the error criterion used for segmentation in Section
\ref{section:one:function}, for instance the $L_2$ . Then a set of functions
is homogeneous if its diameter is small compared to the typical variations of
a function from the set. In the quadratic case, we request for instance to the
maximal quadratic distance between two functions of the set to be small
compared to the variances of functions of the set around their mean values.

Then, if the functions $(s_i)_{i=1}^N$ are assumed to be homogeneous, finding a
single summary function seems natural. In practice, this corresponds to
finding a simple function $g$ that is close to each of the $s_i$ as measured
by a functional norm: one should choose a summary type (e.g., piecewise
constant), a norm (e.g., $L_2$) and a way to aggregate the
individual comparison of each function to the summary (e.g., the sum of the
norms). 

Let us first consider the case of a piecewise constant function $g$ defined by
$P$ intervals $(I_p)_{p=1}^P$ (a partition of $[t_1,t_M]$) and $P$ values
$(a_p)_{p=1}^P$ (with $g(t)=a_p$ when $t\in I_p$). If we measure the distance
between $g$ and $s_i$ via the $L_2$ distances and consider the sum of $L_2$
distances as the target error measure, we obtain the following segmentation
error:
\begin{equation}
  \label{eq:centrality}
E\left((s_i)_{i=1}^N,(I_p)_{p=1}^P,(a_p)_{p=1}^P\right)=\sum_{i=1}^N\sum_{p=1}^P\sum_{t_k\in I_p}\left(s_i(t_k)-a_p\right)^2,
\end{equation}
using the same quadrature scheme as in Section
\ref{subsection:segmentation}. Huygens' theorem gives
\begin{equation}
\sum_{i=1}^N\left(s_i(t_k)-a_p\right)^2=N\left(\mu(t_k)-a_p\right)^2+
\sum_{i=1}^N\left(s_i(t_k)-\mu(t_k)\right)^2,
\end{equation}
where $\mu=\frac{1}{N}\sum_{i=1}^Ns_i$ is the mean function. Therefore
minimizing $E$ from equation \eqref{eq:centrality} is equivalent to minimizing
\begin{equation}
  \label{eq:centrality:mean}
E'\left((s_i)_{i=1}^N,(I_p)_{p=1}^P,(a_p)_{p=1}^P\right)=\sum_{p=1}^P\sum_{t_k\in I_p}\left(\mu(t_k)-a_p\right)^2.
\end{equation}
In other words, the problem consists simply in building an optimal summary of
the mean curve of the set and is solved readily with Algorithm
\ref{algo:dp:single}. The additional computational cost induced by the
calculation of the mean function is in $O(NM)$. This will generally be
negligible compared to the $O(PM^2)$ cost of the dynamic programming. 

\begin{figure}[htbp]
  \centering
\includegraphics[width=0.8\textwidth]{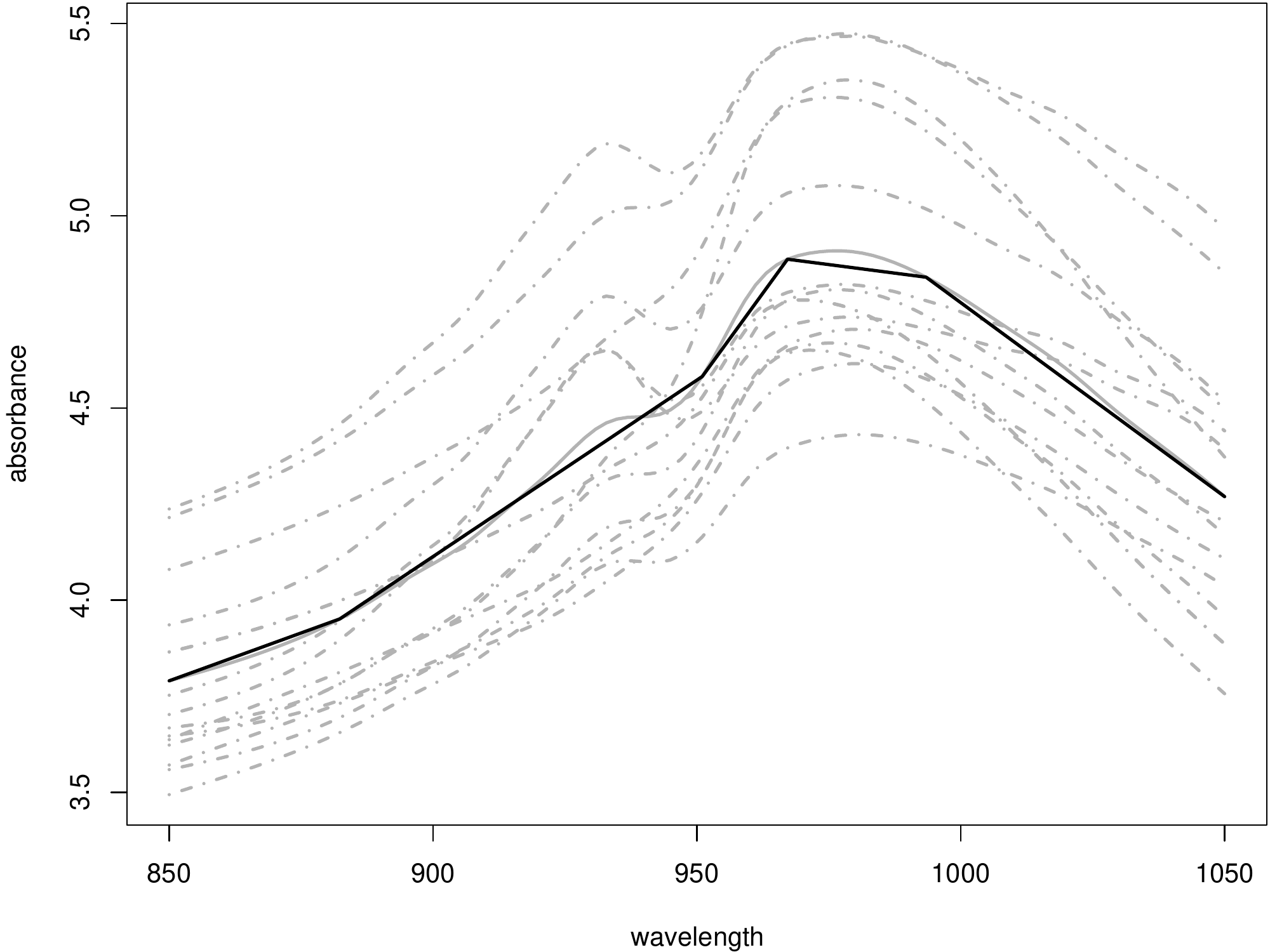}
  \caption{Group summary via an linear interpolation of the mean; the grey
    dashed/dotted curves are the original ones, the grey solid curve is the mean
    curve and the 
    black curve is the piecewise linear interpolation curve that provides the
    group summary}
  \label{fig:Tecator:group:linear}
\end{figure}

The reasoning above applies more generally to piecewise models used with the
sum of $L_2$ distances. Indeed if $g(t)$ is such that $g(t)=g_p(t)$ when
$t\in I_p$, then we have
\begin{equation}
\sum_{i=1}^N\left(s_i(t_k)-g_p(t_k)\right)^2=N\left(\mu(t_k)-g_p(t_k)\right)^2+
\sum_{i=1}^N\left(s_i(t_k)-\mu(t_k)\right)^2,
\end{equation}
which leads to the same solution: $g$ can be obtained as the summary of the
mean curve. The case of linear interpolation proposed in Section
\ref{subsection:extensions} is a bit different in the sense that interpolating
several curves at the same time is not possible. Therefore, the natural way to
extend the proposal to a set of curves it to interpolate the mean curve. An
example of the result obtained by this method is given in Figure
\ref{fig:Tecator:group:linear}. 

In the general case, we assume given an error criterion $Q$ that applies to
a set of contiguous evaluation points $\{t_{k},\ldots,t_{l}\}$: it gives the
error done by the local model on $\{t_{k},\ldots,t_{l}\}$ as an approximation
of all the functions $(s_i)_{i=1}^N$. Then given an ordered partition of
$\{t_{1},\ldots,t_{M}\}$, $(C_p)_{p=1}^P$, we use the sum of the
$Q((s_i)_{i=1}^N,C_p)$ as a global error measure and obtain an additive form
as in equation \eqref{eq:AdditiveCriterion}. Algorithm \ref{algo:dp:single}
applies straightforwardly (it applies also to taking, e.g., the maximum of the
$Q((s_i)_{i=1}^N,C_p)$ over $p$).

The only difficulty compared to the case of a unique function is the
efficient calculation of the $Q((s_i)_{i=1}^N,\{t_{k},\ldots,t_{l}\})$. As
shown above, the $L_2$ case is very favorable as it amounts to summarizing the
mean curve. However, more complex norms and/or aggregation rules lead to
difficulties. If we use for instance
\begin{equation}\label{eq:max:ldeux}
Q((s_i)_{i=1}^N,\{t_{k},\ldots,t_{l}\})=\min_{a}\max_{1\leq i\leq N}\sum_{j=k}^l(s_i(t_j)-a)^2,
\end{equation}
the problem cannot be formulated as a single curve summarizing problem, while
is consists simply in considering the maximal $L_2$ distance rather than the
sum of the $L_2$ distances: if the curves have different variances on
$\{t_{k},\ldots,t_{l}\}$, the optimal value $a$ will not be the mean on the
set of evaluation points of the mean curve. More generally, the computation of
all $Q((s_i)_{i=1}^N,\{t_{k},\ldots,t_{l}\})$ might scale much faster than in
$O(M(M+N))$ and thus might dominate the total computation time. Indeed, if
there is no closed form expression for $Q$, then the value has to be computed
from scratch for each pair $(k,l)$, which implies $O(M^2)$ problem
solving. Then, if no special structure or preprocessing can be used to
simplify the problem, computing one value of $Q$ implies clearly to look at
all the involved functions, i.e, to $O(N)$ observations. Therefore, if no
simplification can be done, computing all values of $Q$ will imply at least at
$O(NM^2)$ calculation cost. Care must therefore be exercised when choosing the
quality criterion and the summarizing model to avoid excessive
costs. That said, the dynamic programming approach applies and leads to an
optimal segmentation according to the chosen criterion.

\subsection{Clustering}\label{secClustering}
Assuming that a set of functions is homogeneous is too strong in practice, as
shown on Figure \ref{fig:Tecator:group:linear}: some of the curves have a
small bump around 940 nm but the bump is very small in the mean curve. As a
consequence, the summary misses this feature while the single curve summary
used in Figure \ref{fig:Tecator:LinVSInterp} was clearly picking up the bump. 

It is therefore natural to rely on a clustering strategy to produce
homogeneous clusters of curves and then to apply the method described in the
previous section to summarize each cluster. However, the direct application of
this strategy might lead to suboptimal solutions: at better solution should be
obtained by optimizing at once the clusters and the associated summaries. We
derive such an algorithm in the present section.

In a similar way to the previous section, we assume given an error measure
$Q(G,\{t_{k},\ldots,t_{l}\})$ (where $G$ is a subset of $\{1,\ldots,N\}$): it
measures the error made by the local model on 
the $\{t_{k},\ldots,t_{l}\}$ as an approximation of the functions $s_i$ for
$i\in G$. For instance
\begin{equation}\label{eq:quality:constant:sum:ldeux}
Q(G,\{t_{k},\ldots,t_{l}\})=\min_{a}\sum_{i\in G}\sum_{j=k}^l(s_i(t_j)-a)^2,
\end{equation}
for a constant local model evaluated with respect to the sum of $L_2$
distances. Let us assume for now that each cluster is summarized via a
segmentation scheme that uses a cluster independent number of segments, $P/K$
(this value is assumed to be an integer). Then, given a partition of
$\{1,\ldots,N\}$ into $K$ clusters $(G_k)_{k=1}^K$ and given for each cluster
an ordered partition of $\{t_{1},\ldots,t_{M}\}$, $(C^k_p)_{p=1}^{P/K}$, the
global error made by the summarizing scheme is
\begin{equation}
  \label{eq:globalquality}
E\left((s_i)_{i=1}^N,(G_k)_{k=1}^K,\left((C^k_p)_{p=1}^{P/K}\right)_{k=1}^K\right)=\sum_{k=1}^K\sum_{p=1}^{P/K}Q(G_k,C^k_p).
\end{equation}
Building an optimal summary of the set of curves amounts to minimizing this
global error with respect to the function partition and, inside each cluster,
with respect to the segmentation. 

A natural way to tackle the problem is to apply an alternating minimization
scheme (as used for instance in the K-means algorithm): we optimize
successively the partition given the segmentation and the segmentation given
the partition. Given the function partition $(G_k)_{k=1}^K$, the error
is a sum of independent terms: one can apply dynamic programming to build an
optimal summary of each of the obtained clusters, as shown in the previous
section. Optimizing with respect to the partition given the segmentation is a
bit more complex as the feasibility of the solution depends on the actual
choice of $Q$. In general however, $Q$ is built by aggregating via the maximum
or the sum operators distances between the functions assigned to a cluster and
the corresponding summary. It is therefore safe to assume that assigning a
function to the cluster with the closest summary in term of the distance used
to define $Q$ will give the optimal partition given the summaries. This leads
to Algorithm \ref{algo:kmeans:uniform}.
\begin{algorithm}
\caption{Simultaneous clustering and segmentation algorithm (uniform case)}
\label{algo:kmeans:uniform}
  \begin{algorithmic}[1]
\STATE initialize the partition $(G_k)_{k=1}^K$ randomly
\REPEAT
  \FOR{$k=1$ to $K$}
    \STATE apply Algorithm \ref{algo:dp:single} to the find the optimal
    summary $g_k$ of the set of curves $\{s_i|i\in G_k\}$ in $P/K$ segments
  \ENDFOR
  \FOR{$i=1$ to $N$}
     \STATE assign $s_i$ to $G_k$ such that the distance between $s_i$ and
     $g_k$ is minimal over $k$
  \ENDFOR
\UNTIL{$(G_k)_{k=1}^K$ is stable}
  \end{algorithmic}
\end{algorithm}
The computational cost depends strongly on the availability of an efficient
calculation scheme for $Q$. If this is the case, then the cost of the $K$
dynamic programming applications grows in $O(K(P/K)M^2)=O(PM^2)$. With
standard $L_p$ distances between functions, the cost of the assignment phase
is in $O(NKM)$. Computing the mean function in each cluster (or similar
quantities such as the median function that might be needed to compute
efficiently $Q$) has a negligible cost compared to the other costs (e.g.,
$O(NM)$ for the mean functions). 

Convergence of Algorithm \ref{algo:kmeans:uniform} is obvious as long as both
the assignment phase (i.e., the minimization with respect to the partition
given the segmentation) and the dynamic programming algorithm use
deterministic tie breaking rules and lead to unique solutions. While the issue
of ties between prototypes is generally not relevant in the K-means algorithm,
for instance, it plays an important role in our context. Indeed, the summary
of a cluster can be very simple and therefore, several summaries might have
identical distance to a given curve. Then if ties are broken at random, the
algorithm might never stabilize. Algorithm \ref{algo:dp:single} is formulated
to break ties deterministically by favoring short segments at the beginning of
the segmentation (this is implied by the strict inequality test used at line
10 of the Algorithm). Then the optimal segmentation is unique given the
partition, and the optimal partition is unique given the segmentation. As a
consequence, the alternative minimization scheme converges.

\begin{figure}[htbp]
  \centering
\includegraphics[width=\textwidth]{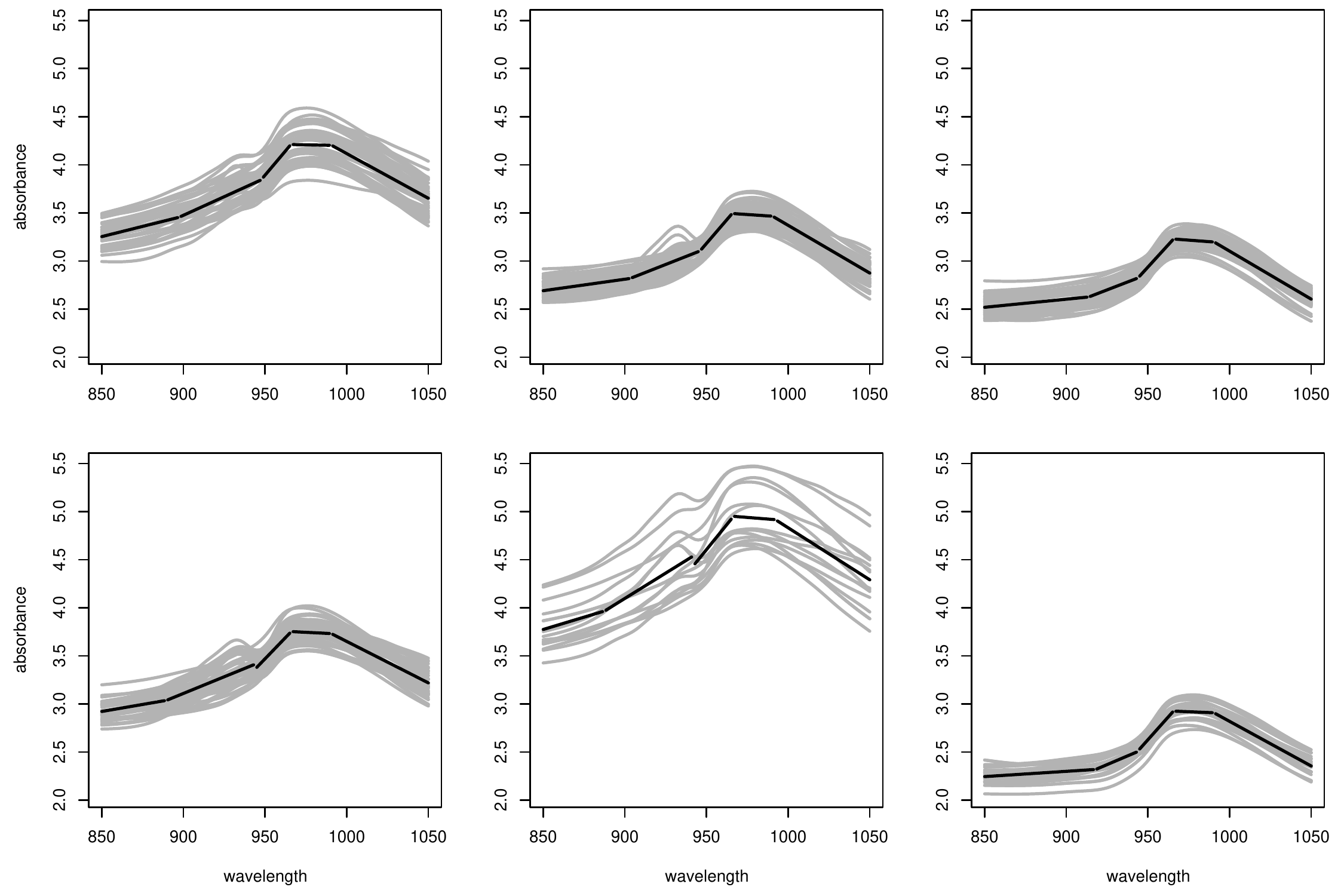}
  \caption{Summary of the full Tecator dataset with 6 clusters and 5 segments
    per cluster: grey curves are the original spectra, dark curves are the
    summaries for each cluster}
  \label{fig:Tecator:Full}
\end{figure}

Figure \ref{fig:Tecator:Full} gives an example of the results obtained by the
algorithm in the case of a piecewise linear summary with five segments per
cluster (the full Tecator dataset contains $N=240$ spectra with $M=100$ for
each spectrum).

In practice, we will frequently use a piecewise constant model whose quality
is assessed via the sum of the $L_2$ distances, as in equation
\eqref{eq:quality:constant:sum:ldeux}. Then equation \eqref{eq:globalquality}
can be instantiated as follows:
\begin{equation}
  \label{eq:constrained:kmeans}
E\left((s_i)_{i=1}^N,(G_k)_{k=1}^K,\left((C^k_p)_{p=1}^{P/K}\right)_{k=1}^K\right)=
\sum_{k=1}^K\sum_{i\in G_k}\sum_{p=1}^{P/K}\sum_{t_j\in C^k_p}\left(s_i(t_j)-\mu_{k,p}\right)^2,
\end{equation}
with
\begin{equation}
\mu_{k,p}=\frac{1}{|G_k||C^k_p|}\sum_{i\in G_k}\sum_{t_j\in C^k_p}s_i(t_j).
\end{equation}
Let us denote $\mathbf{s}_i=(s_i(t_1),\ldots,s_i(t_M))$ the $\R{M}$ vector
representation of the sampled functions and
$\mathbf{g}_k=(g_k(t_1),\ldots,g_k(t_M))$  the vector representation of the
piecewise constant functions defined by $g_k(t_i)=\mu_{k,p}$ when $t_i\in
C^k_p$. Then equation \eqref{eq:constrained:kmeans} can be rewritten
\begin{equation}
  \label{eq:constrained:kmeans:vector}
E\left((\mathbf{s}_i)_{i=1}^N,(G_k)_{k=1}^K,\left((C^k_p)_{p=1}^P\right)_{k=1}^K\right)= \sum_{k=1}^K\sum_{i\in G_k}\|\mathbf{s}_i-\mathbf{g}_k\|^2.
\end{equation}
This formulation emphasizes the link between the proposed approach and
the K means algorithm. Indeed the criterion is the one optimized by the K
means algorithm for the vector data $\mathbf{s}_i$
with a constraint on the prototypes: rather than allowing arbitrary prototypes
as in the standard K means, we use simple prototypes in the sense of Section
\ref{subsection:simple:function}. Any piecewise model whose quality
is assessed via the sum of the $L_2$ distances leads to the same formulation,
but with different constraints on the shape of the functional prototypes. 

Equation \eqref{eq:constrained:kmeans:vector} opens the way to many variations
on the same theme: most of the prototype based clustering methods can be
modified to embed constraints on the prototypes, as Algorithm
\ref{algo:kmeans:uniform} does for the K means. Interesting candidates for
such a generalization includes the Self Organizing Map in its batch version
\cite{KohonenSOM1995}, Neural Gas also in its batch version
\cite{CottrellEtAlBatchNeuralGas2006NN} and deterministic annealing based
clustering \cite{rosedeterministicannealing1999},
among others. 

\subsection{Optimal resources allocation}
A drawback of Algorithm \ref{algo:kmeans:uniform} is that it does not allocate
resources in an optimal way: we assumed that each summary will use $P/K$
segments, regardless of the cluster. Fortunately,
dynamic programming can be used again to remove this constraint. Let us
consider indeed a resource assignment, that is $K$ positive integers
$(P_k)_{k=1}^K$ such that $\sum_{k=1}^KP_k=P$. The error made by the
summarizing scheme when cluster $G_k$ uses $P_k$ segments for its summary is
given by
\begin{equation}
  \label{eq:globalquality:assign}
E\left((s_i)_{i=1}^N,(G_k)_{k=1}^K,(P_k)_{k=1}^K,\left((C^k_p)_{p=1}^{P_k}\right)_{k=1}^K\right)=\sum_{k=1}^K\sum_{p=1}^{P_k}Q(G_k,C^k_p). 
\end{equation}
As for the criterion given by equation \eqref{eq:globalquality} in the
previous section, we rely on an alternating minimization scheme. Given the
function partition $(G_k)_{k=1}^K$, the challenge is to optimize $E$ with
respect to both the resource assignment and the segmentation. For a
fixed partition $(G_k)_{k=1}^K$, let us denote 
\begin{equation}
  \label{eq:globalquality:subcost}
R_k(P_k)=\min_{(C^k_p)_{p=1}^{P_k}\text{ ordered partition of }\{t_1,\ldots,t_M\}}\sum_{p=1}^{P_k}Q(G_k,C^k_p)
\end{equation}
Then minimizing $E$ from equation \eqref{eq:globalquality:assign}
for a fixed partition $(G_k)_{k=1}^K$ corresponds to minimizing:
\begin{equation}
  \label{eq:globalquality:optassign}
U\left((P_k)_{k=1}^K\right)=\sum_{k=1}^KR_k(P_k)
\end{equation}
with respect to the resource assignment. This formulation emphasizes two major
points. 

Firstly, as already mentioned in Section \ref{secOptimalSegmentation}, given a
number of segments $P$ Algorithm \ref{algo:dp:single} and its variants provide
with a negligible additional cost all the optimal segmentations in $p=1$ to
$P$ segments. Therefore, the calculation of $R_k(p)$ for
$p\in\{1,\ldots,P\}$ can be done with a variant of Algorithm
\ref{algo:dp:single} in $O(PM^2)$ provided an efficient calculation of
$Q(G_k,C^k_p)$ is possible (in fact, one needs only values of $R_k(p)$ for
$p\leq P-K+1$ as we request at least one segment per cluster, but to ease the
calculation we assume that we compute values up to $P$). 

Secondly, $U$ as defined in equation \eqref{eq:globalquality:optassign} is an
additive measure and can therefore be optimized by dynamic programming. Indeed
let us define $S(l,p)$ as 
\begin{equation}
  \label{eq:globalquality:subcost:dp}
S(l,p) =\min_{(P_k)_{k=1}^l\text{ such that }\sum_{k=1}^lP_k=p}\sum_{k=1}^lR_k(P_k)
\end{equation}
The $S(1,p)$ are readily obtained from the optimal segmentation algorithm
($S(1,p)=R_1(p)$). Then the additive structure of equation
\eqref{eq:globalquality:optassign} shows that
\begin{equation}
  \label{eq:globalquality:subcost:dp:rec}
S(l,p)=\min_{1\leq u\leq p-l+1}S(l-1,p-u)+R_l(u).
\end{equation}
Given all the $R_l(p)$ the calculation of $S(K,P)$ has therefore a cost of
$O(KP^2)$. This calculation is plugged into Algorithm
\ref{algo:kmeans:uniform} to obtain Algorithm \ref{algo:kmeans:optimal}. 

\begin{algorithm}
\caption{Simultaneous clustering and segmentation algorithm with optimal
  resource assignment}
\label{algo:kmeans:optimal}
  \begin{algorithmic}[1]
\STATE initialize the partition $(G_k)_{k=1}^K$ randomly
\REPEAT
  \FOR{$k=1$ to $K$}
    \STATE apply Algorithm \ref{algo:dp:single} to compute $R_k(p)$ for all
    $p\in\{1,\ldots,P-K+1\}$ 
  \ENDFOR
  \FOR[initialisation of the resource allocation phase]{$p=1$ to $P-K+1$}
     \STATE $S(1,p)\leftarrow R_1(p)$
  \ENDFOR
  \FOR[dynamic programming for resource allocation]{$l=2$ to $K$}
     \FOR{$p=l$ to $P$}
        \STATE $W(l,p)\leftarrow 1$
        \STATE $S(l,p)\leftarrow S(l-1,p-1)+R_l(1)$
        \FOR{$u=2$ to $p-l+1$}
          \IF{$S(l-1,p-u)+R_l(u)<S(l,p)$}
            \STATE $W(l,p)\leftarrow u$
            \STATE $S(l,p)\leftarrow S(l-1,p-u)+R_l(u)$
          \ENDIF
        \ENDFOR
     \ENDFOR
  \ENDFOR
  \STATE $P_{opt}\leftarrow (\text{NA},\ldots,\text{NA})$
  \COMMENT{backtracking phase}
  \STATE $P_{opt}(K)\leftarrow W(K,P)$
  \STATE $availabe \leftarrow P-W(K,P)$
  \FOR{$k=K-1$ to $1$}
    \STATE $P_{opt}(k)\leftarrow W(k,available)$
    \STATE $availabe \leftarrow availabe - W(k,available)$
  \ENDFOR
  \FOR[summary calculation]{$k=1$ to $K$}
     \STATE build $g_k$ with $P_{opt}(k)$ segments 
  \ENDFOR
  \FOR{$i=1$ to $N$}
     \STATE assign $s_i$ to $G_k$ such that the distance between $s_i$ and
     $g_k$ is minimal over $k$
  \ENDFOR
\UNTIL{$(G_k)_{k=1}^K$ is stable}
  \end{algorithmic}
\end{algorithm}
Given the values of $Q$, the total computational cost of the
optimal resource assignment and of the optimal segmentation is therefore in
$O(KP(M^2+P))$. As we target simple descriptions, it seems reasonable to assume
that $P\ll M^2$ and therefore that the cost is dominated by $O(KPM^2)$. Then,
using the optimal resource assignment multiplies by $K$ the computational cost
compared to a uniform use of $P/K$ segments per cluster
(which computational cost is in $O(PM^2)$). The main source of additional
computation is the fact that one has to study summaries using $P$ segments for
each cluster rather than only $P/K$ in the uniform case. This cost can be
reduced by introducing an arbitrary (user chosen) maximal number of segments
in each clusters, e.g., $\lambda P/K$. Then the running time of the optimal
assignment is $\lambda$ times the running time of the uniform solution. As we
aim at summarizing the data, using a small $\lambda$ such $2$ seems
reasonable. 

The computation of the values of $Q$ remains the same regardless of the way
the number of segments is chosen for each cluster. The optimisation of the
error with respect to the partition  $(G_k)_{k=1}^K$ follows also the same
algorithm in both cases. If we use for instance piecewise constant
approximation evaluated with the sum of the $L_2$ distances, then, as already
mentioned previously, the summaries are computed from the cluster means:
computing the means costs $O(NM)$, then the
calculation of $Q$ for all functional clusters is in $O(KM^2)$  and the
optimization with respect to $(G_k)_{k=1}^K$ is in $O(NKM)$. In general, the
dominating cost will therefore scale in $O(KPM^2)$. However, as pointed out in
Section \ref{subsection:single:summary}, complex choices for the $Q$ error
criterion may induce larger computational costs. 

\subsection{Two phases approaches}\label{subsection:two:phases}
The main motivation of the introduction of Algorithm \ref{algo:kmeans:uniform}
was to avoid relying on a simpler but possibly suboptimal dual phases
solution. This alternative solution simply consists applying a standard K
means to the vector representation of the functions $(s_i)_{i=1}^N$ to get
homogeneous clusters and then to summarize each cluster via a simple
prototype. While this approach should intuitively give higher errors for the
criterion of equation \eqref{eq:globalquality}, the actual situation is
more complex.

Algorithms \ref{algo:kmeans:uniform} and \ref{algo:kmeans:optimal} are both
optimal in their segmentation phase and will generally be also optimal for the
partition determination phase if $Q$ is a standard criterion. However, the
alternating optimization scheme itself is not optimal (the K means algorithm
is also non optimal for the same reason). We have therefore no guarantee on
the quality of the local optimum reached by those algorithms, exactly as in
the case of the K means. In addition, even if the two phase approach is
initialized with the same random partition as, e.g., Algorithm
\ref{algo:kmeans:uniform}, the final partitions can be distinct because of the
distinct prototypes used during the iterations. It is therefore difficult to
predict which approach is more likely to give the best solution. In addition a
third possibility should be considered: the K means algorithm is applied to
the vector representation of the functions $(s_i)_{i=1}^N$ to get homogeneous
clusters and is followed by an application of Algorithm
\ref{algo:kmeans:uniform} or \ref{algo:kmeans:optimal} using the obtained
clusters for its initial configuration.

To assess the differences between these three possibilities, we conducted a
simple experiment: using the full Tecator dataset, we compared the value of
$E$ from equation \eqref{eq:globalquality} for the final summary obtained by
the three solutions using the same random starting partitions with $K=6$,
$P=30$ and piecewise constant summaries. We used 50 random initial
configurations and compared the methods in the case of uniform resource
allocation ($5$ segments per cluster) and in the case of optimal resource
allocation. 

In the case of uniform resources assignment the picture is rather
simple. For all the 50 random initialisations both two phases algorithms gave
exactly the same results. In 44 cases, using  Algorithm
\ref{algo:kmeans:uniform} gave exactly the same results as the two phases
algorithms. In 5 cases, its results were of lower quality and in the last
case, its results were slightly better. The three algorithms manage to reach
the same overall optimum (for which $E=472$). 

The case of optimal resources assignment is a bit more complex. For 15 out of
50 initial configurations, using Algorithm \ref{algo:kmeans:optimal} after the
K means improves the results over a simple optimal summary (with optimal
resource assignment) applied to the results of the K means (results were
identical for the 35 other cases). In 34 cases, K means followed by Algorithm
\ref{algo:kmeans:optimal} gave the same results as Algorithm
\ref{algo:kmeans:optimal} alone. In the 16 remaining cases, results are in
favor of K means followed by Algorithm \ref{algo:kmeans:optimal} which wins 13
times. Nevertheless, the three algorithms manage to reach
the same overall optimum (for which $E=467$). 

While there is no clear winner in this (limited scope) experimental analysis,
considering the computational cost of the three algorithms helps choosing the
most appropriate one in practice. In favorable cases, the cost of an iteration
of Algorithm \ref{algo:kmeans:optimal} is $O(KMN+KPM^2))$ while the cost of an
iteration of the K means on the same dataset is $O(KMN)$. In functional data,
$M$ is frequently of the same order as $N$. For summarizing purposes, $P$
should remain small, around $5K$ for instance. This means that one can perform
several iterations of the standard K means for the same computational price
than one iteration of Algorithm \ref{algo:kmeans:optimal}. For instance, the
Tecator dataset values, $P=30$, $M=100$ and $N=240$, give $1+PM/N=13.5$. In
the experiments described above, the median number of iterations for the K
means was 15 while it was respectively 15 and 16 for Algorithms
\ref{algo:kmeans:uniform} and \ref{algo:kmeans:optimal}. We can safely assume
therefore that Algorithm \ref{algo:kmeans:optimal} is at least 10 times slower
than the K means on the Tecator dataset. However, when initialized with the K
means solution, Algorithms \ref{algo:kmeans:uniform} and
\ref{algo:kmeans:optimal} converges very quickly, generally in two iterations
(up to five for the optimal assignment algorithm in the Tecator
dataset). In more complex situations, as the ones studied in Section
\ref{secExperiments} the number of iterations is larger but remains small
compared to what would be needed when starting from a random configuration. 

Based on the previous analysis, we consider that using the K means algorithm
followed by Algorithm \ref{algo:kmeans:optimal} is the best solution. If
enough computing resources are available, one can run the K means algorithm
from several starting configurations, keep the best final partition and use it
as an initialization point for Algorithm \ref{algo:kmeans:optimal}. As the K
means provides a good starting point to this latter algorithm, the number of
iterations remains small and the overall complexity is acceptable, especially
compared to using Algorithm \ref{algo:kmeans:optimal} alone from different
random initial configurations. Alternatively, the analyst can rely on more
robust clustering scheme such as Neural Gas 
\cite{CottrellEtAlBatchNeuralGas2006NN} and deterministic annealing based
clustering \cite{rosedeterministicannealing1999}, or on the Self Organizing
Map \cite{KohonenSOM1995} to provide the starting configuration of Algorithm
\ref{algo:kmeans:optimal}. 

\section{Experimental results}\label{secExperiments}
We give in this section two examples of the type of exploratory analyses that
can be performed on real world datasets with the proposed method. In order to
provide the most readable display for the user, we build the initial
configuration of Algorithm \ref{algo:kmeans:optimal} with a batch Self
Organizing Map. We optimize the initial radius of the neighborhood influence
in the SOM with the help of the Kaski and Lagus topology preservation
criterion \cite{KaskiLagusICANN1996} (the radius is chosen among 20 radii). 

\subsection{Topex/Poseidon satellite}
\begin{figure}[htbp]
  \centering
\includegraphics[width=0.9\textwidth]{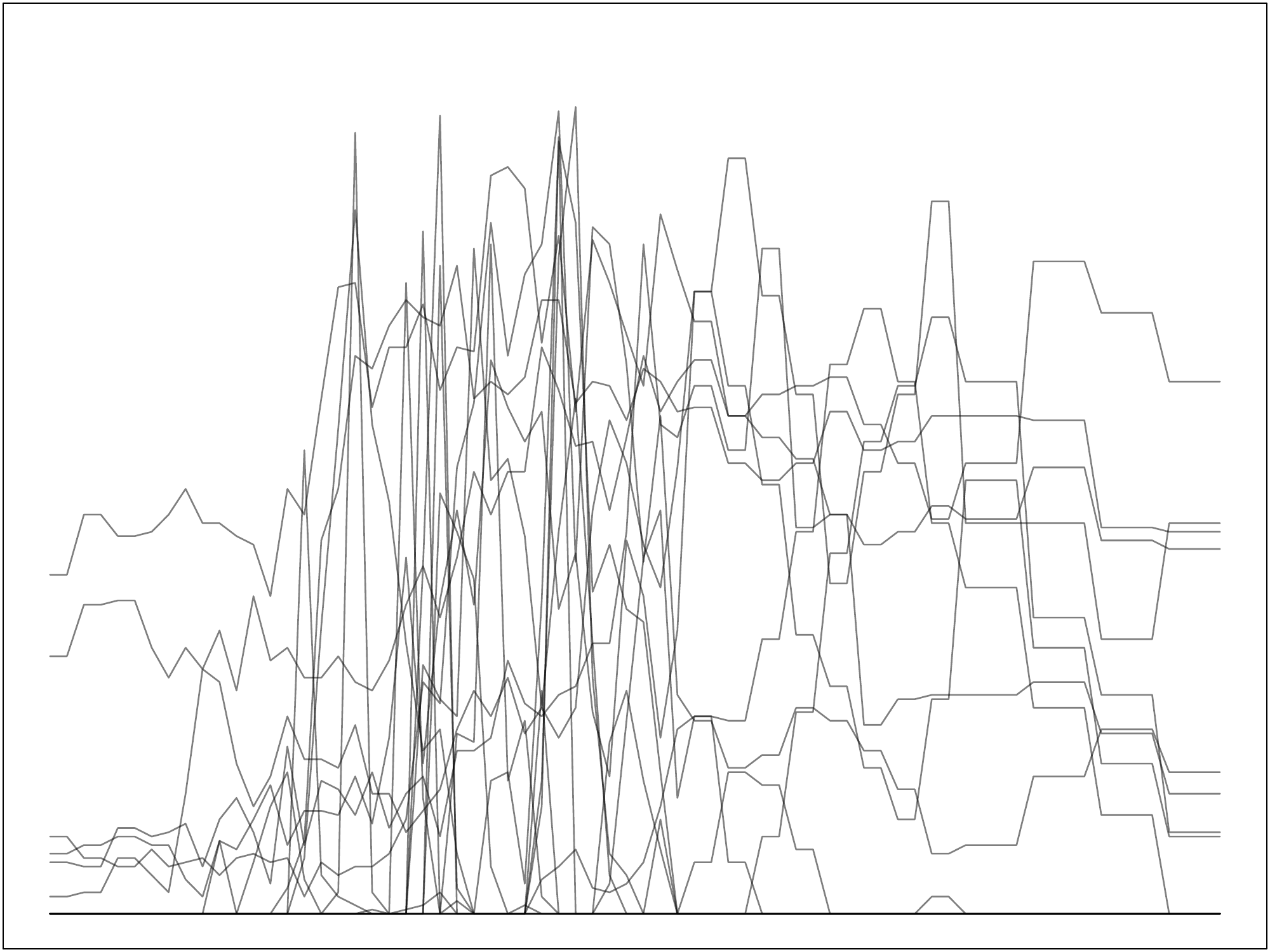}  
  \caption{20 curves from the Topex/Poseidon radar satellite dataset}
  \label{fig:Topex:curves}
\end{figure}

We study first the Topex/Poseidon satellite dataset\footnote{Data are
  available at
  \url{http://www.lsp.ups-tlse.fr/staph/npfda/npfda-datasets.html}}.  The
Topex/Poseidon radar satellite has been used over the Amazonian basin to
produce $N=472$ waveforms sampled at $M=70$ points. The curves exhibit
a quite large variability induced by differences in ground type below the
satellite during data acquisition. Details on the dataset can be found in
e.g. \cite{Frappart2003,Frappart2005}. Figure \ref{fig:Topex:curves} displays
20 curves from the dataset chosen randomly. 

\begin{figure}[htbp]
  \centering
\includegraphics[width=0.9\textwidth]{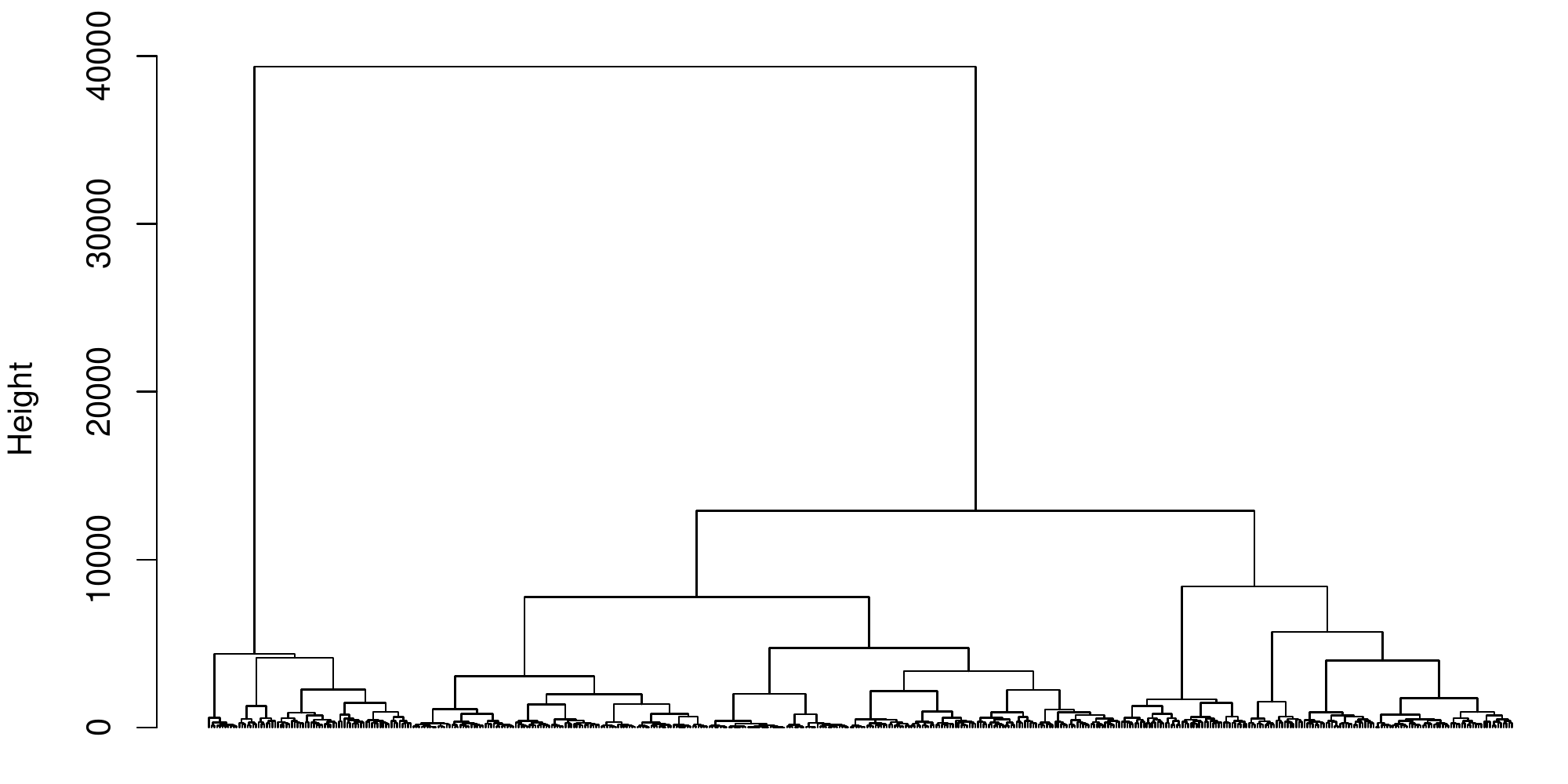}  
  \caption{Dendrogram of the hierarchical clustering for the Topex/Poseidon radar satellite dataset}
  \label{fig:Topex:hclust}
\end{figure}

\begin{figure}[htbp]
  \centering
\includegraphics[width=0.9\textwidth]{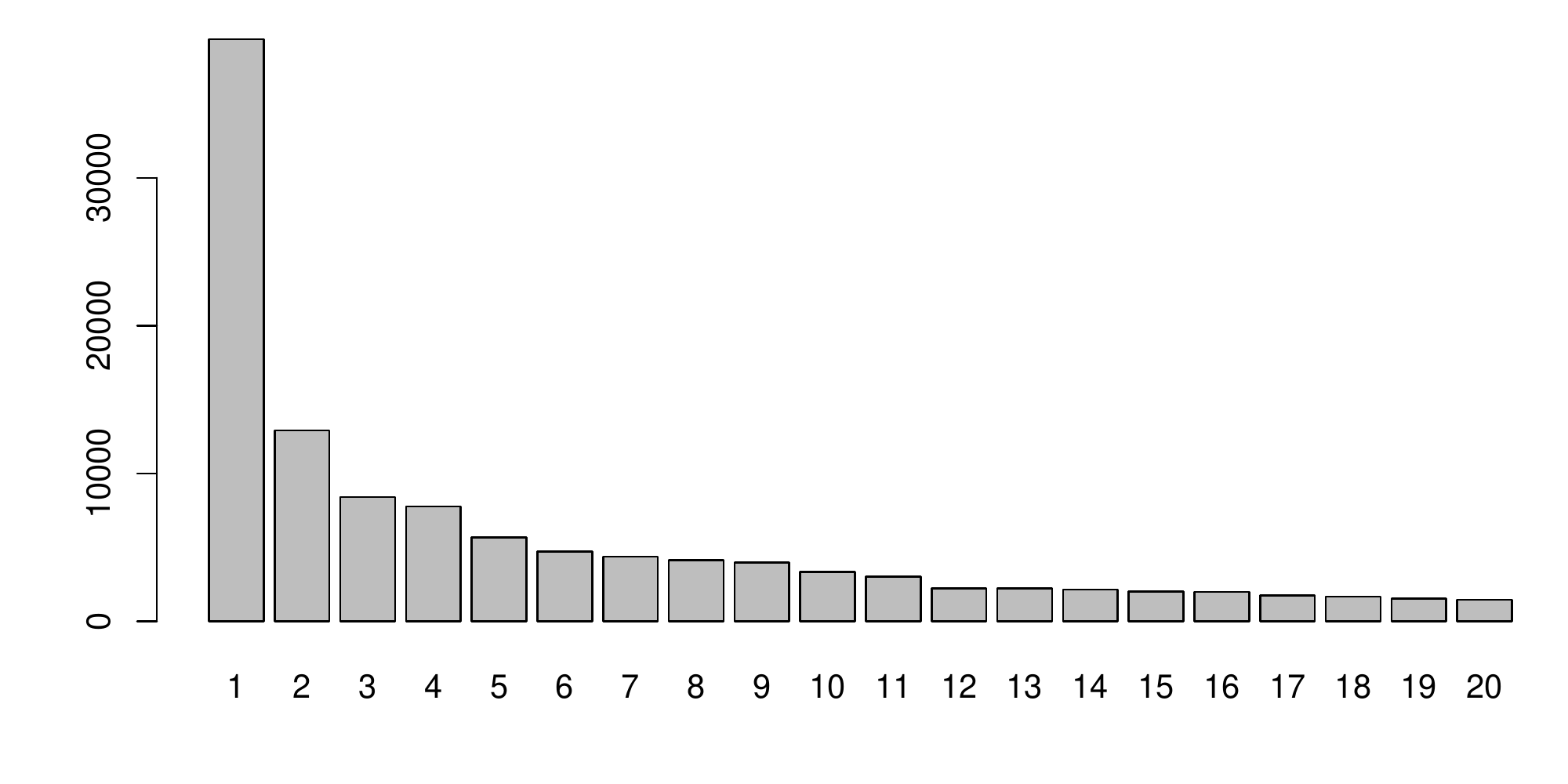}  
  \caption{Total within class variance decrease for the 20 first merges in the
  dendrogram depicted by Figure \ref{fig:Topex:hclust}}
  \label{fig:Topex:hclust:heights}
\end{figure}

We conduct first a hierarchical clustering (based on Euclidean distance
between the functions and the Ward criterion) to get a rough idea of the
potential number of clusters in the dataset. Both the dendrogram (Figure
\ref{fig:Topex:hclust}) and the total
within class variance evolution (Figure \ref{fig:Topex:hclust:heights}) tend to
indicate a rather small number of clusters, around 12. As the visual
representation obtained via the SOM is more readable than a randomly arranged
set of prototypes, we can use a slightly oversized SOM without impairing the
analyst's work. We decided therefore to use a $4\times 5$ grid: the rectangular
shape has been preferred over a square one according to results from
\cite{DBLP:conf/esann/UltschH05} that show some superiority in topology
preservation for anisotropic maps compared to isotropic ones. 

\begin{figure}[htbp]
  \centering
\includegraphics[width=\textwidth]{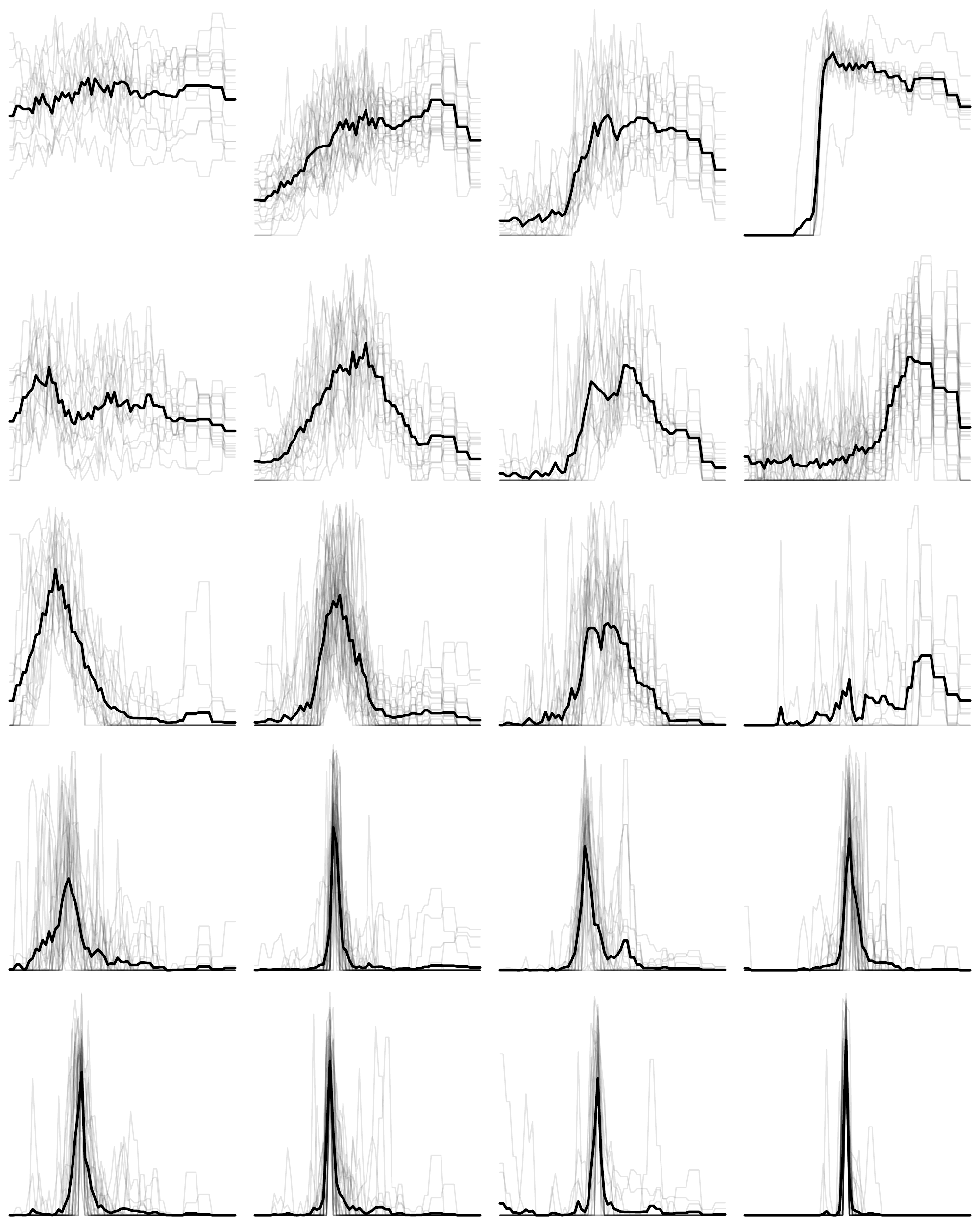}  
  \caption{Prototypes obtained by SOM algorithm on the Topex/Poseidon radar
    satellite dataset (grey curves are the original curves as assigned to the
    clusters)}
  \label{fig:Topex:SOM}
\end{figure}

The results of the SOM are given in Figure \ref{fig:Topex:SOM} that displays
the prototype of each cluster arranged on the SOM grid. Each cell contains
also all the curves assigned to the associated clusters. As expected, the
SOM prototypes are well organized on the grid: the horizontal axis
seems to encode approximately the position of the maximum of the prototype
while the vertical axis corresponds to the width of the peak. However, the
prototypes are very noisy and inferring the general behavior of the curves of
a cluster remains difficult. 
\begin{figure}[htbp]
  \centering
\includegraphics[width=\textwidth]{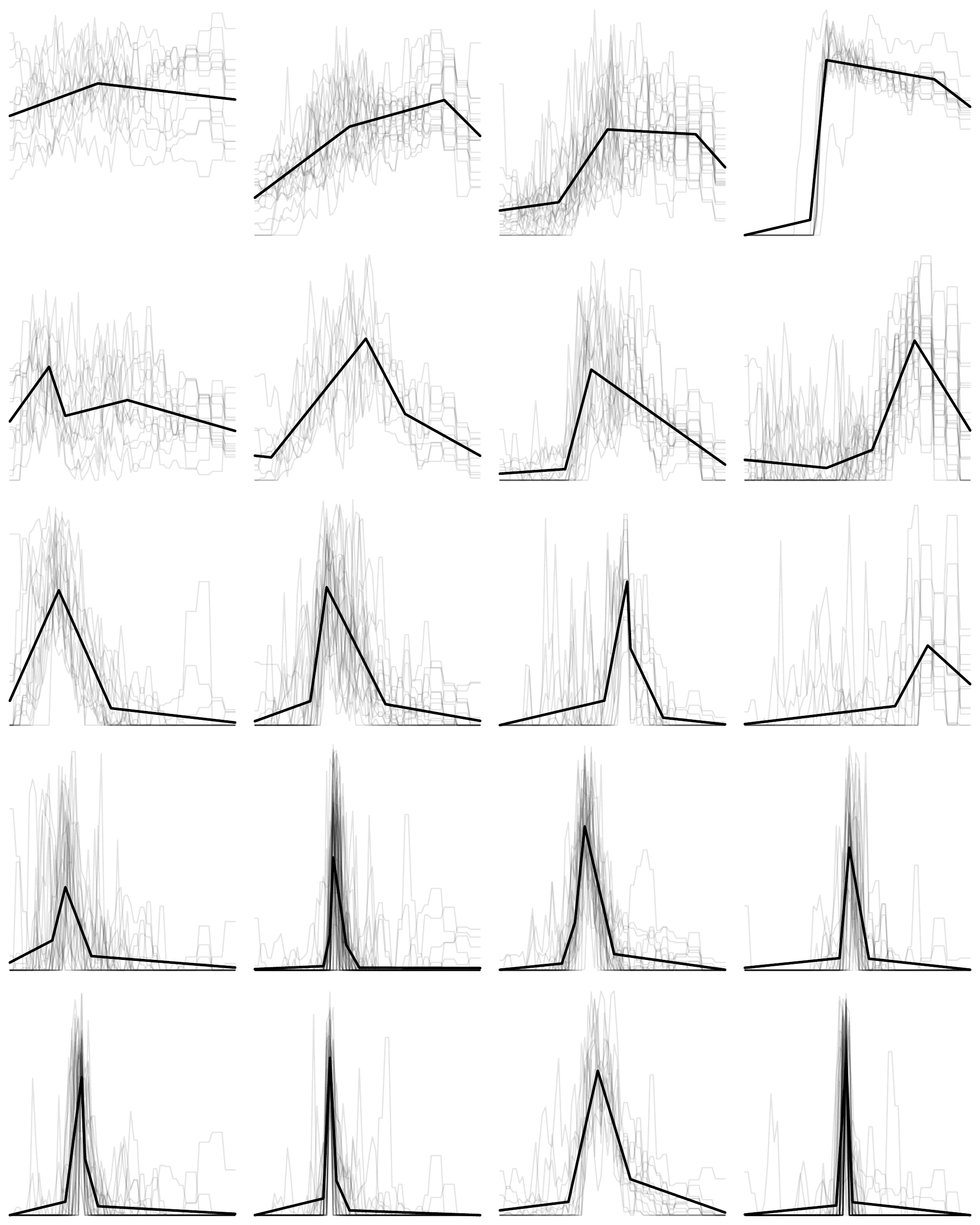}  
  \caption{Summarized prototypes obtained by Algorithm \ref{algo:kmeans:optimal}  on the Topex/Poseidon radar
    satellite dataset (grey curves are the original curves as assigned to the
    clusters)}
  \label{fig:Topex:hclust:SOM:simple}
\end{figure}

Figure \ref{fig:Topex:hclust:SOM:simple} represents the results obtained after
applying Algorithm \ref{algo:kmeans:optimal} to the results of the SOM. In
order to obtain a readable summary, we set $P$ to $80$, i.e., an average of 4
segments for each cluster. We use the linear interpolation approach described
in Section \ref{subsection:extensions}. Contrarily to results obtained on the
Tecator dataset and reported in Section \ref{subsection:two:phases}, Algorithm
\ref{algo:kmeans:optimal} used a significant number of iterations (17). Once
the new clusters and the simplified prototypes were obtained, they were
displayed in the same way as for the SOM results. The resulting map is
almost as well organized as the original one (with the exception of a large
peak on the bottom line), but the simplified prototypes are much more
informative than the original ones: they give an immediate idea of the general
behavior of the curves in the corresponding cluster. For example, the
differences between the two central prototypes of the second line (starting
from the top) are more obvious with summaries: in the cluster on the left, one
can identify a slow increase followed by a two phases slow decrease, while the
other cluster corresponds to a sharp increase followed by a slow
decrease. Those differences are not as obvious on the original map.

\subsection{Electric power consumption}
\begin{figure}[htbp]
  \centering
\includegraphics[width=0.9\textwidth]{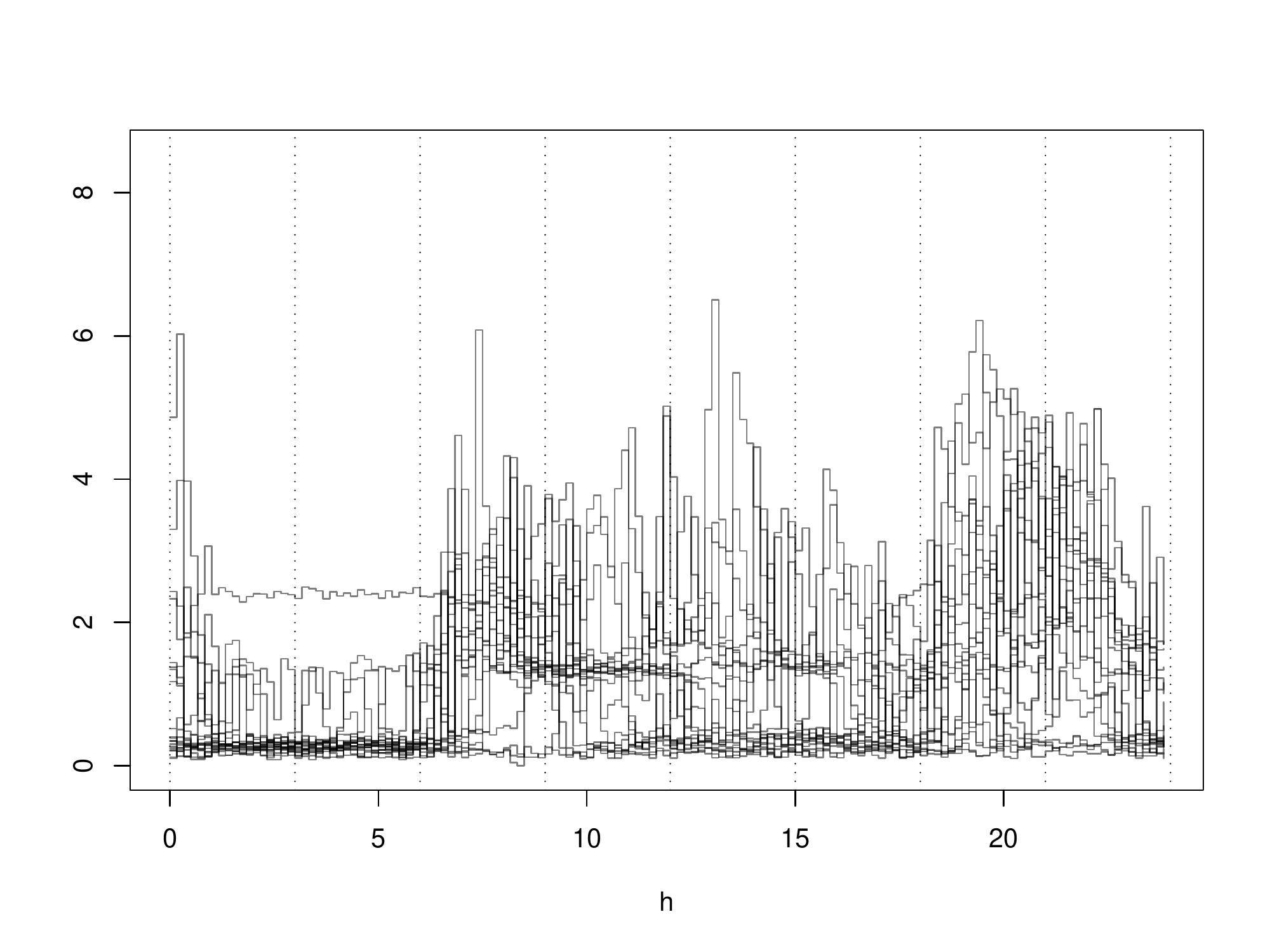}  
  \caption{20 electric power consumption curves}
  \label{fig:Conso:curves}
\end{figure}

The second dataset\footnote{This dataset is available at
  \url{http://bilab.enst.fr/wakka.php?wiki=HomeLoadCurve}.} consists in the
electric power consumption recorded in a personal home during almost one year
(349 days). Each curve consists in 144 measures which give the power
consumption of one day at a 10 minutes sampling rate. Figure
\ref{fig:Conso:curves} displays 20 randomly chosen curves.

\begin{figure}[htbp]
  \centering
\includegraphics[width=0.9\textwidth]{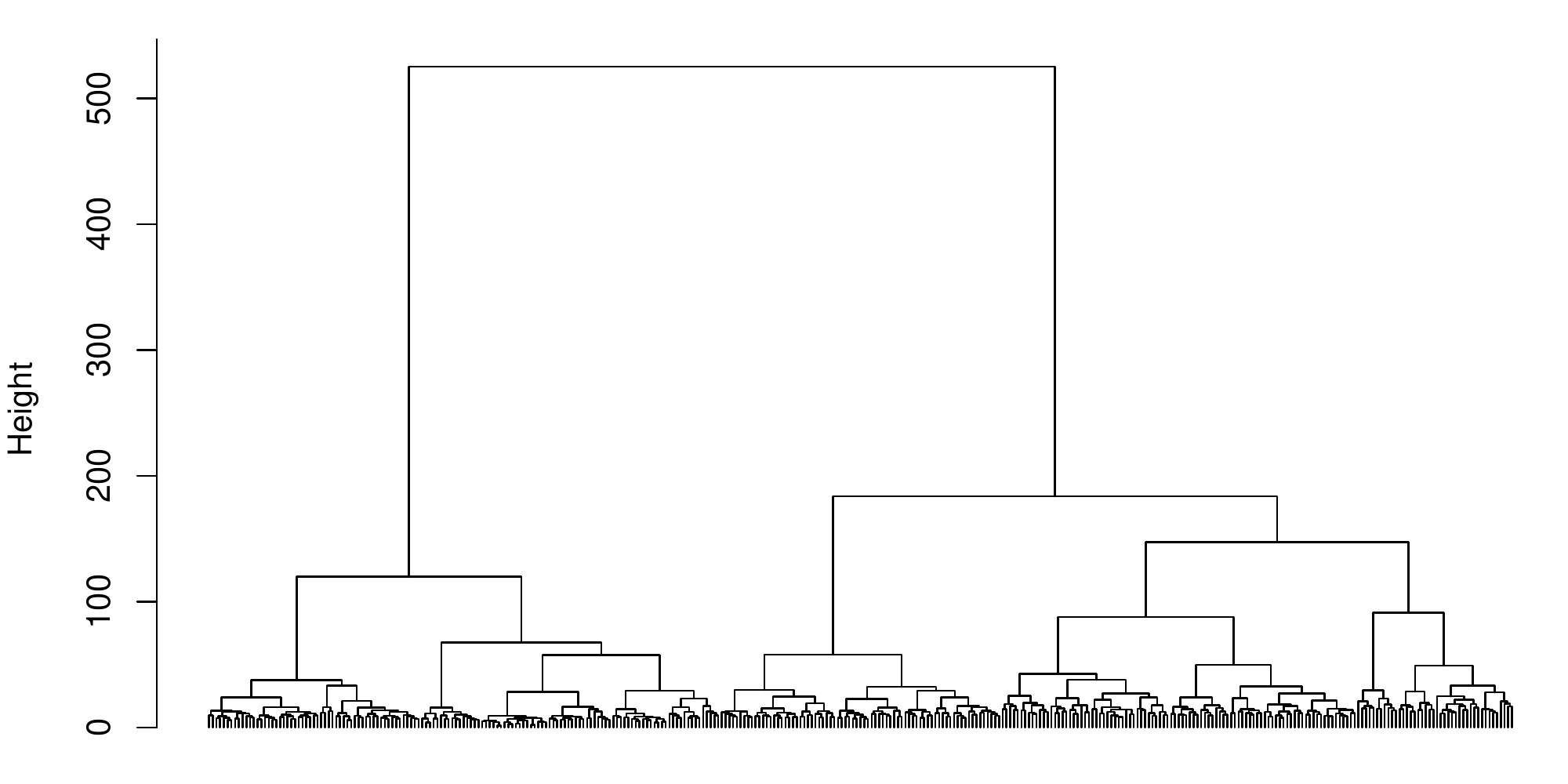}  
  \caption{Dendrogram of the hierarchical clustering for the load curves}
  \label{fig:Conso:hclust}
\end{figure}

\begin{figure}[htbp]
  \centering
\includegraphics[width=0.9\textwidth]{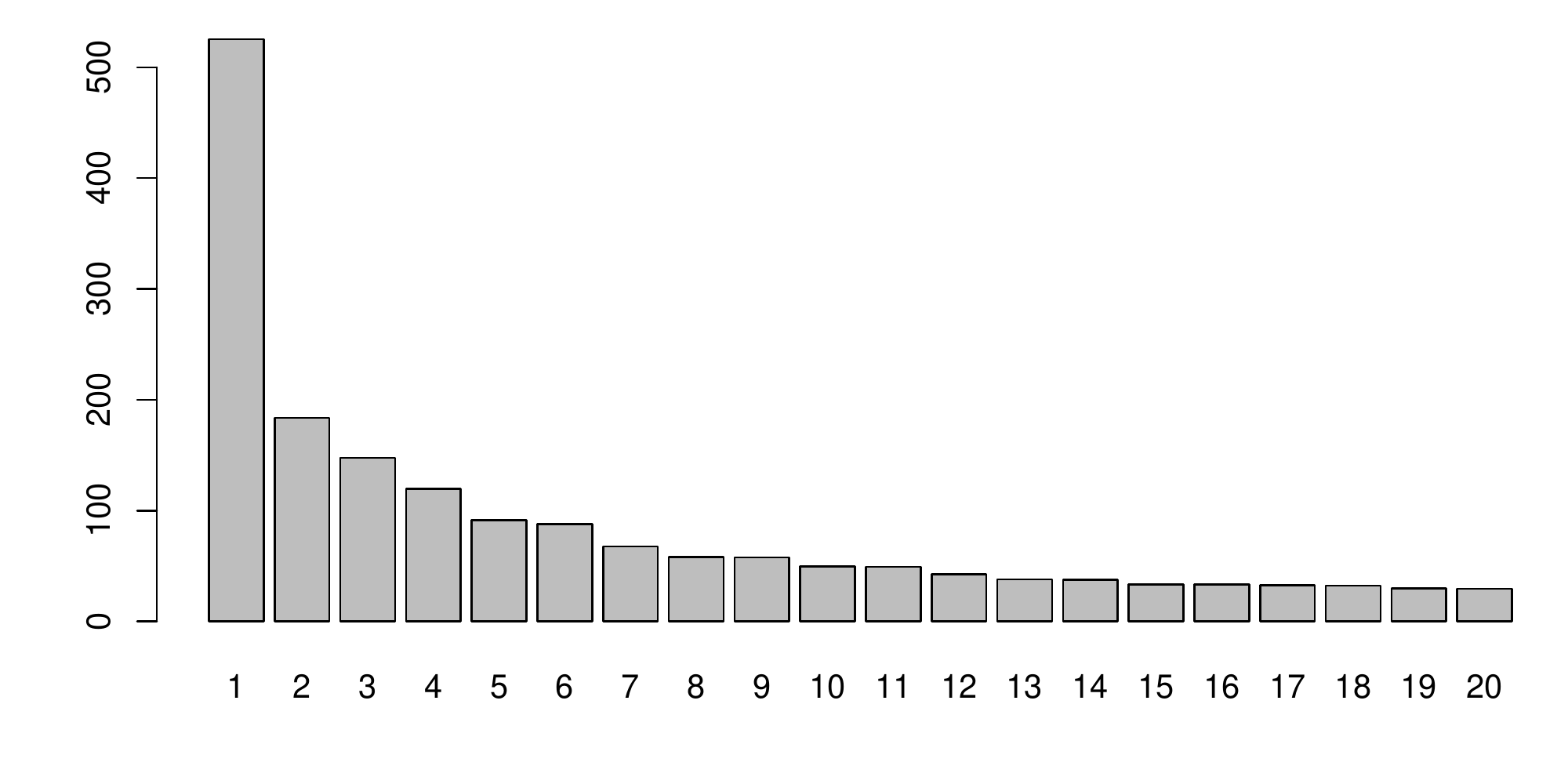}  
  \caption{Total within class variance decrease for the 20 first merges in the
  dendrogram depicted by Figure \ref{fig:Conso:hclust}}
  \label{fig:Conso:hclust:heights}
\end{figure}

\begin{figure}[htbp]
  \centering
\includegraphics[width=\textwidth]{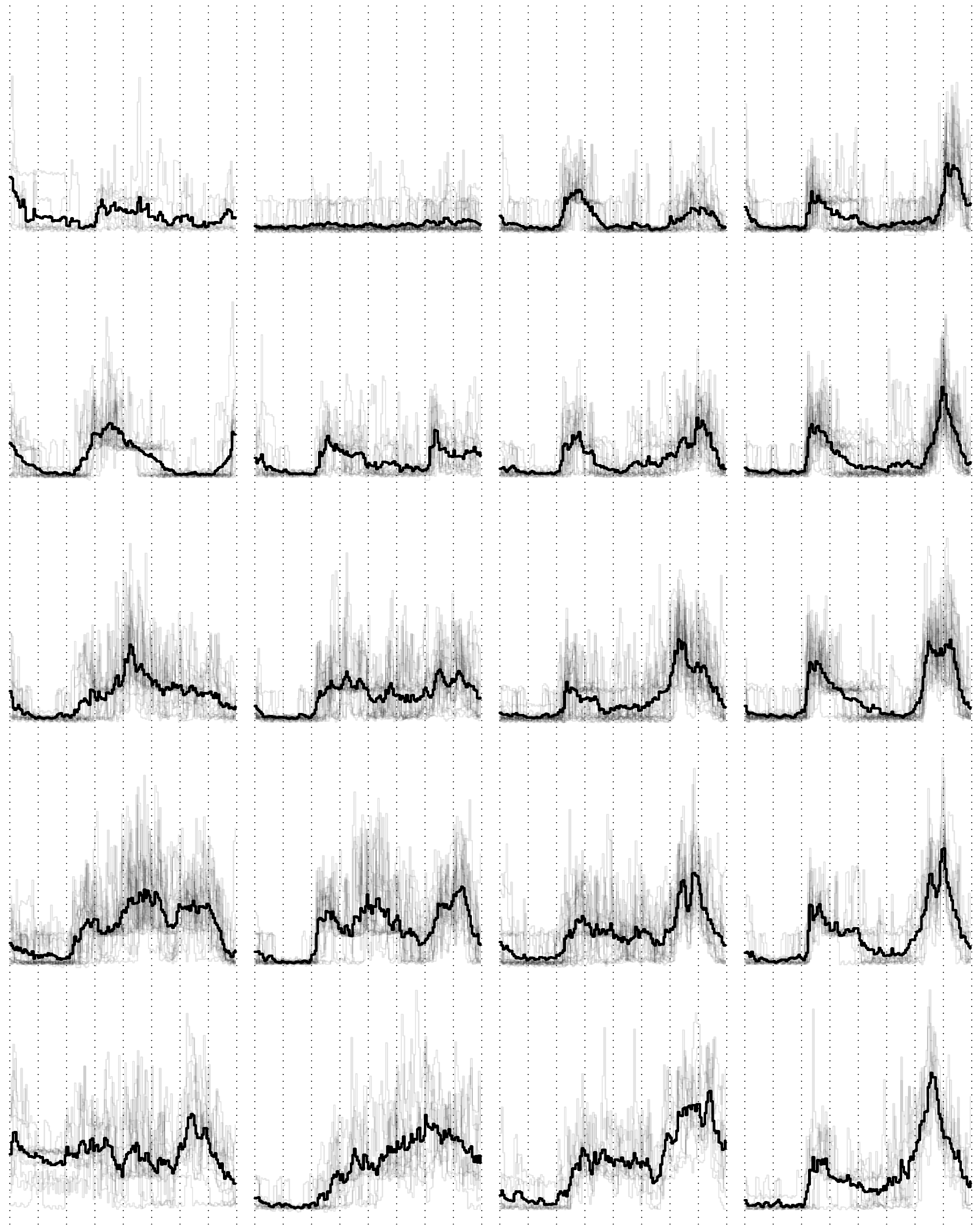}  
  \caption{Prototypes obtained by SOM algorithm for the load curves (grey curves are the original curves as assigned to the
    clusters)}
  \label{fig:Conso:SOM}
\end{figure}

We analyse the curves in a very similar way as for the Topex/Poseidon
dataset The results of the hierarchical clustering displayed by Figures
\ref{fig:Conso:hclust} and \ref{fig:Conso:hclust:heights} lead to the same
conclusion as in the previous case: the number of clusters seems to be small,
around 13. As before, we fit a slightly larger SOM ($4\times 5$ rectangular
grid) to the data and obtain the representation provided by Figure
\ref{fig:Conso:SOM}. The map is well organized: the vertical axis seems to
encode the global power consumption while the horizontal axis represents
the overall shape of the load curve. However, the shapes of the prototypes are
again not easy to interpret, mainly because of their noisy nature. 

\begin{figure}[htbp]
  \centering
\includegraphics[width=\textwidth]{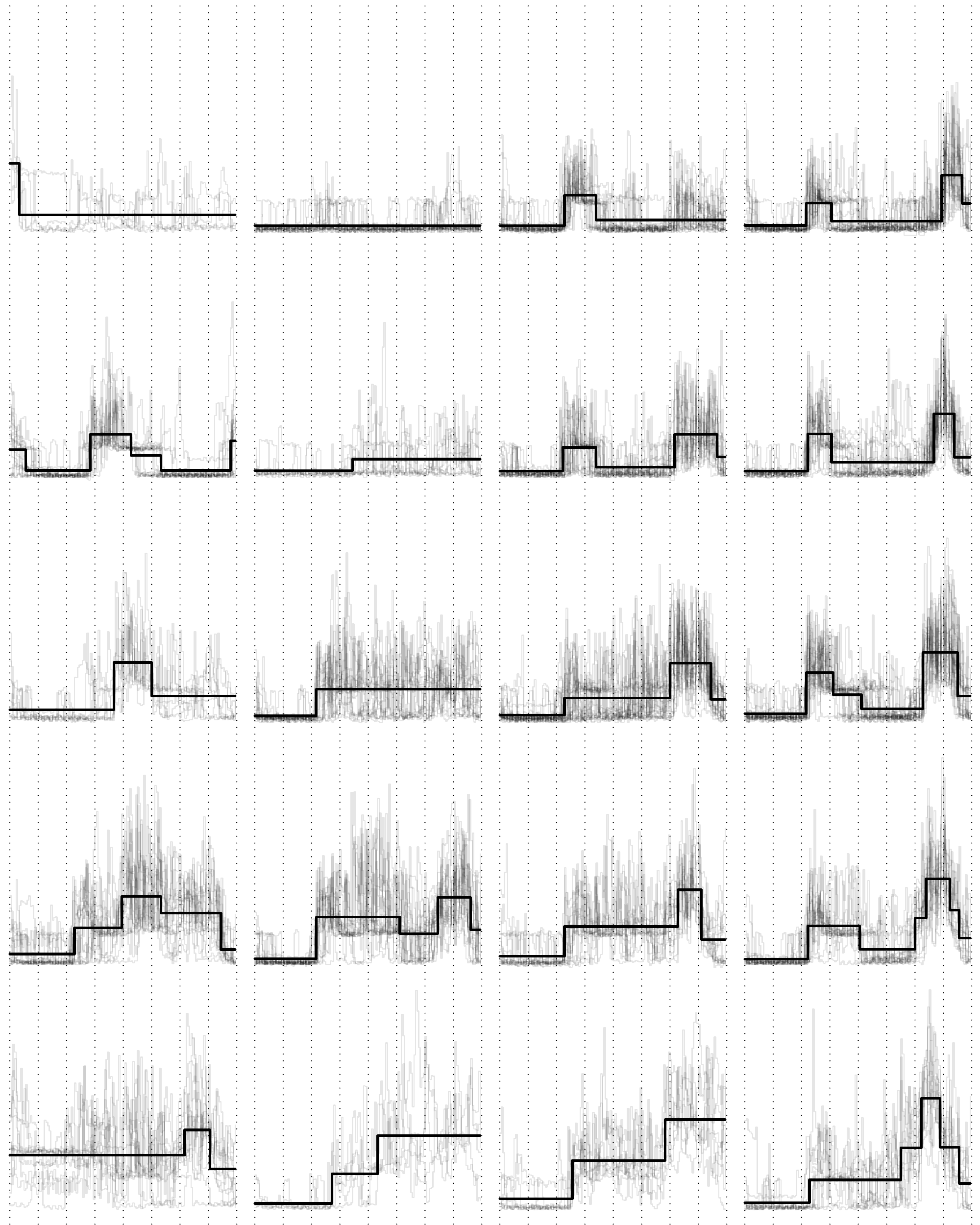}  
  \caption{Summarized prototypes ($P=80$) obtained by Algorithm
    \ref{algo:kmeans:optimal}  for the load curves (grey curves are the original curves as assigned to the
    clusters)}
  \label{fig:Conso:hclust:SOM:simple}
\end{figure}

Figure \ref{fig:Conso:hclust:SOM:simple} shows the results obtained after
applying Algorithm \ref{algo:kmeans:optimal} to the results of the SOM
(Algorithm \ref{algo:kmeans:optimal} used 9 iterations to reach a stable
state). In order to obtain a readable summary, we set $P$ to $80$, i.e., an
average of 4 segments for each cluster. We depart from the previous analysis
by using here a piecewise constant summary: this is more adapted to load
curves as they should be fairly constant on significant time periods, because
many electrical appliances have stable power consumption once switched on. As
for the Topex/Poseidon dataset, the summarized prototypes are much easier to
analyze than their noisy counterpart. Even the global organization of the map
is more obvious: the top right of the map for instance, gathers days in which
the power consumption in the morning (starting at 7 am) is significant, is
followed by a period of low consumption and then again by a consumption peak
starting approximately at 7 pm. Those days are week days in which the home
remains empty during work hours. Other typical clusters include holidays with
almost no consumption (second cluster on the first line), more constant days
which correspond to week ends, etc. In addition, the optimal resource
assignment tends to emphasize the difference between the clusters by
allocating less resources to simple ones (as expected). This is quite obvious
in the case of piecewise constant summaries used in Figure
\ref{fig:Conso:hclust:SOM:simple} (and to a lesser extent on Figure
\ref{fig:Topex:hclust:SOM:simple}). 

\begin{figure}[htbp]
  \centering
\includegraphics[width=\textwidth]{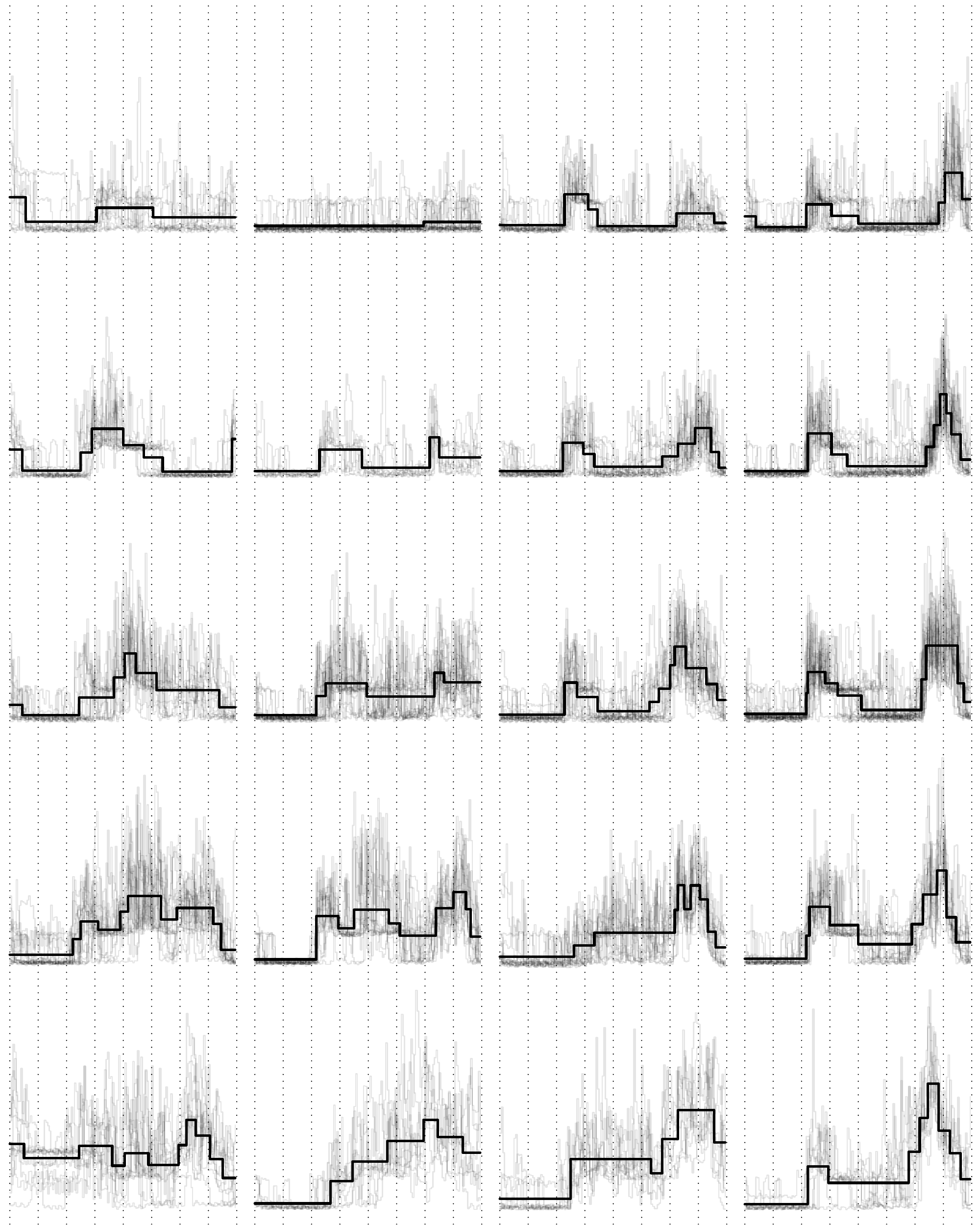}  
  \caption{Summarized prototypes ($P=160$) obtained by Algorithm
    \ref{algo:kmeans:optimal}  for the load curves (grey curves are the original curves as assigned to the
    clusters)}
  \label{fig:Conso:hclust:SOM:simple:double}
\end{figure}

In order to emphasize the importance of a drastic reduction in prototype
complexity, Algorithm \ref{algo:kmeans:optimal} was applied to the same
dataset, using the same initial configuration, with $P=160$, i.e., an average
of 8 segments in each cluster. Results are represented on Figure
\ref{fig:Conso:hclust:SOM:simple:double}. 
As expected, prototypes are more complex than on Figure
\ref{fig:Conso:hclust:SOM:simple}. While they are not as rough as the original
ones (see Figure \ref{fig:Conso:SOM}), their analysis is more difficult than
the one of prototypes obtained with stronger summarizing constraints. In
particular, some consumption peaks are no more roughly outlined as in Figure
\ref{fig:Conso:hclust:SOM:simple} but more accurately approximated (see for
instance the last cluster of the second row). This leads to series of short
segments from which expert inference is not easy. It should be noted in
addition, that because of the noisy nature of the data, increasing the number
of segments improves only marginally the approximation quality. To show this,
we first introduce a relative approximation error measure given by
\begin{equation}
  \label{eq:relativeerror}
\frac{\sum_{k=1}^K\sum_{i\in G_k}\sum_{p=1}^{P_k}\sum_{t_j\in C^k_p}(s_i(t_j)-\mu_{k,p})^2}{\sum_{i=1}^N\sum_{j=1}^M(s_i(t_j)-\mu_i)^2},
\end{equation}
using notations from Section \ref{secClustering}, and where $\mu_i$ denotes
the mean of $s_i$. This relative error compares the total squared
approximation error to the total squared internal variability inside the
dataset.  Table \ref{tab:Conso:errors} gives the dependencies between the
number of segments and the relative approximation error. It is quite clear
that the relative loss in approximation precision is acceptable even with
strong summarizing constraints. 
\begin{table}[hbtp]
  \centering
  \begin{tabular}{lccc}
Summary & No summary & $P=160$ & $P=80$\\\hline
Relative approximation error &  0.676 & 0.696 & 0.740  \\\hline
  \end{tabular}
  \caption{Approximation errors for the load curves}
  \label{tab:Conso:errors}
\end{table}
This type of numerical quality assessment should be provided to the analyst
either as a way to control the relevance of the summarized prototypes, or as a
selection guide for choosing the value of $P$. 

\section{Conclusion}
We have proposed in this paper a new exploratory analysis method for
functional data. This method relies on simple approximations of functions,
that is on models that are piecewise constant or piecewise linear. Given a set
of homogeneous functions, a dynamic programming approach computes efficiently
and optimally a simple model via a segmentation of the interval on which the
functions are evaluated. Given several groups of homogeneous functions and a
total number of segments, another dynamic programming algorithm allocates
optimally to each cluster the number of segments used to build the
summary. Finally, those two algorithms can be embedded in any classical
prototype based algorithm (such as the K means) to refine the groups of
functions given the optimal summaries and \emph{vice versa} in an alternating
optimization scheme.

The general framework is extremely flexible and accommodates different
types of quality measures ($L_2$, $L_1$), aggregation strategies (maximal
error, mean error), approximation model (piecewise constant, linear
interpolation) and clustering strategies (K means, Neural gas). Experiments
show that it provides meaningful summaries that help greatly the analyst in
getting quickly a good idea of the general behavior of a set of curves. 

While we have listed many variants of the method to illustrate its
flexibility, numerous were not mentioned as they are more distant from the
main topic of this paper. It should be noted for instance that rather than
building a unique \emph{summary} for a set of curves, one could look for a
unique \emph{segmentation} supporting curve by curve summaries. In other
words, each curve will be summarized by e.g., a piecewise constant model, but
all models will share the same segmentation. This strategy is very interesting
for analyzing curves in which internal changes are more important than the
actual values taken by the curves. It is closely related to the so called best
basis problem in which a functional basis is built to represent efficiently a
set of curves (see e.g.,
\cite{CoifmanWickerhauser1992BestBasis,SaitoCoifman1995LocalBases,RossiLechevallier2008SFC}). This
is also related to variable clustering methods (see e.g.,
\cite{KrierEtAl2007CILSFDProj}). This final link opens an interesting
perspective for the proposed method. It has been shown recently
\cite{FrancoisEtAl2008Agrostat,KrierEtAlESANN2009} that variable clustering
methods can be adapted to supervised learning: variables are grouped in a way
that preserves as much as possible the explanatory power of the reduced set of
variables with respect to a target variable. It would be extremely useful to
provide the analyst with such a tool for exploring a set of curves: she would
be able to select an explanatory variable and to obtain summaries of those
curves that mask details that are not relevant for predicting the chosen
variable.  

\section*{Acknowledgment}
The authors thank the anonymous referees for their valuable comments that
helped improving this paper. 

\bibliographystyle{abbrvnat}
\bibliography{func-clust-seg}

\begin{thebibliography}{34}
\providecommand{\natexlab}[1]{#1}
\providecommand{\url}[1]{\texttt{#1}}
\expandafter\ifx\csname urlstyle\endcsname\relax
  \providecommand{\doi}[1]{doi: #1}\else
  \providecommand{\doi}{doi: \begingroup \urlstyle{rm}\Url}\fi

\bibitem[Abraham et~al.(2003)Abraham, Cornillon, Matzner-L\/ober, and
  Molinari]{Abraham2000}
C.~Abraham, P.-A. Cornillon, E.~Matzner-L\/ober, and N.~Molinari.
\newblock Unsupervised curve clustering using b-splines.
\newblock \emph{Scandinavian Journal of Statistics}, 30\penalty0 (3):\penalty0
  581--595, September 2003.

\bibitem[Auger and Lawrence(1989)]{AugerLawrence1989}
I.~E. Auger and C.~E. Lawrence.
\newblock Algorithms for the optimal identification of segment neighborhoods.
\newblock \emph{Bulletin of Mathematical Biology}, 51\penalty0 (1):\penalty0
  39--54, 1989.

\bibitem[Bellman(1961)]{Bellman1961Function}
R.~Bellman.
\newblock On the approximation of curves by line segments using dynamic
  programming.
\newblock \emph{Communication of the ACM}, 4\penalty0 (6):\penalty0 284, 1961.

\bibitem[Chamroukhi et~al.(2009)Chamroukhi, Sam{\'e}, Govaert, and
  Aknin]{ChamroukhiEtAlESANN2009}
F.~Chamroukhi, A.~Sam{\'e}, G.~Govaert, and P.~Aknin.
\newblock A regression model with a hidden logistic process for signal
  parametrization.
\newblock In \emph{Proceedings of XVIth European Symposium on Artificial Neural
  Networks (ESANN 2009)}, pages 503--508, Bruges, Belgique, April 2009.

\bibitem[Coifman and Wickerhauser(1992)]{CoifmanWickerhauser1992BestBasis}
R.~R. Coifman and M.~V. Wickerhauser.
\newblock Entropy-based algorithms for best basis selection.
\newblock \emph{IEEE Transactions on Information Theory}, 38\penalty0
  (2):\penalty0 713--718, March 1992.

\bibitem[Cottrell et~al.(1998)Cottrell, Girard, and
  Rousset]{CottrellGirardRousset1998}
M.~Cottrell, B.~Girard, and P.~Rousset.
\newblock Forecasting of curves using a kohonen classification.
\newblock \emph{Journal of Forecasting}, 17:\penalty0 429--439, 1998.

\bibitem[Cottrell et~al.(2006)Cottrell, Hammer, Hasenfu\ss, and
  Villmann]{CottrellEtAlBatchNeuralGas2006NN}
M.~Cottrell, B.~Hammer, A.~Hasenfu\ss, and T.~Villmann.
\newblock Batch and median neural gas.
\newblock \emph{Neural Networks}, 19\penalty0 (6--7):\penalty0 762--771,
  July--August 2006.

\bibitem[Debr{\'e}geas and H{\'e}brail(1998)]{DebregeasHebrail1998Curves}
A.~Debr{\'e}geas and G.~H{\'e}brail.
\newblock Interactive interpretation of kohonen maps applied to curves.
\newblock In \emph{Proc. International Conference on Knowledge Discovery and
  Data Mining (KDD'98)}, pages 179--183, New York, August 1998.

\bibitem[Devroye et~al.(1996)Devroye, Gy\"orfi, and
  Lugosi]{DevroyeEtAl1996Pattern}
L.~Devroye, L.~Gy\"orfi, and G.~Lugosi.
\newblock \emph{A Probabilistic Theory of Pattern Recognition}, volume~21 of
  \emph{Applications of Mathematics}.
\newblock Springer, 1996.

\bibitem[Fran{\c c}ois et~al.(2008)Fran{\c c}ois, Krier, Rossi, and
  Verleysen]{FrancoisEtAl2008Agrostat}
D.~Fran{\c c}ois, C.~Krier, F.~Rossi, and M.~Verleysen.
\newblock Estimation de redondance pour le clustering de variables spectrales.
\newblock In \emph{Actes des 10{\`e}mes journ{\'e}es Europ{\'e}ennes
  Agro-industrie et M{\'e}thodes statistiques (Agrostat 2008)}, pages 55--61,
  Louvain-la-Neuve, Belgique, January 2008.

\bibitem[Frappart(2003)]{Frappart2003}
F.~Frappart.
\newblock Catalogue des formes d'onde de l'altim{\`e}tre topex/pos{\'e}idon sur
  le bassin amazonien.
\newblock Rapport technique, CNES, Toulouse, France, 2003.

\bibitem[Frappart et~al.(2006)Frappart, Calmant, Cauhop{\'e}, Seyler, and
  Cazenave]{Frappart2005}
F.~Frappart, S.~Calmant, M.~Cauhop{\'e}, F.~Seyler, and A.~Cazenave.
\newblock Preliminary results of envisat ra-2-derived water levels validation
  over the amazon basin.
\newblock \emph{Remote Sensing of Environment}, 100\penalty0 (2):\penalty0 252
  -- 264, 2006.
\newblock ISSN 0034-4257.

\bibitem[Healey et~al.(1995)Healey, Booth, and
  Enns]{HealeyEtal1995Preattentive}
C.~G. Healey, K.~S. Booth, and J.~T. Enns.
\newblock Visualizing real-time multivariate data using preattentive
  processing.
\newblock \emph{ACM Transactions on Modeling and Computer Simulation},
  5\penalty0 (3):\penalty0 190--221, July 1995.

\bibitem[Huber(1964)]{HuberLoss1964}
P.~J. Huber.
\newblock Robust estimation of a location parameter.
\newblock \emph{Annals of Mathematical Statistics}, 35\penalty0 (1):\penalty0
  73--101, 1964.

\bibitem[Hugueney(2006)]{HugueneyPKDD2006}
B.~Hugueney.
\newblock Adaptive segmentation-based symbolic representations of time series
  for better modeling and lower bounding distance measures.
\newblock In \emph{Proceedings of 10th European Conference on Principles and
  Practice of Knowledge Discovery in Databases, PKDD 2006}, volume 4213 of
  \emph{Lecture Notes in Computer Science}, pages 545--552, September 2006.

\bibitem[Jackson et~al.(2005)Jackson, Scargle, Barnes, Arabhi, Alt, Gioumousis,
  Gwin, Sangtrakulcharoen, Tan, and Tsai]{JacksonEtAl2005}
B.~Jackson, J.~Scargle, D.~Barnes, S.~Arabhi, A.~Alt, P.~Gioumousis, E.~Gwin,
  P.~Sangtrakulcharoen, L.~Tan, and T.~T. Tsai.
\newblock An algorithm for optimal partitioning of data on an interval.
\newblock \emph{IEEE Signal Processing Letters}, 12\penalty0 (2):\penalty0
  105--108, Feb. 2005.

\bibitem[Kaski and Lagus(1996)]{KaskiLagusICANN1996}
S.~Kaski and K.~Lagus.
\newblock Comparing self-organizing maps.
\newblock In C.~von~der Malsburg, W.~von Seelen, J.~Vorbr{\"u}ggen, and
  B.~Sendhoff, editors, \emph{Proceedings of International Conference on
  Artificial Neural Networks (ICANN'96)}, volume 1112 of \emph{Lecture Notes in
  Computer Science}, pages 809--814. Springer, Berlin, Germany,, July 16-19
  1996.

\bibitem[Kohonen(1995)]{KohonenSOM1995}
T.~Kohonen.
\newblock \emph{Self-Organizing Maps}, volume~30 of \emph{Springer Series in
  Information Sciences}.
\newblock Springer, third edition, 1995.
\newblock Last edition published in 2001.

\bibitem[Krier et~al.(2008)Krier, Rossi, Fran{\c c}ois, and
  Verleysen]{KrierEtAl2007CILSFDProj}
C.~Krier, F.~Rossi, D.~Fran{\c c}ois, and M.~Verleysen.
\newblock A data-driven functional projection approach for the selection of
  feature ranges in spectra with ica or cluster analysis.
\newblock \emph{Chemometrics and Intelligent Laboratory Systems}, 91\penalty0
  (1):\penalty0 43--53, March 2008.

\bibitem[Krier et~al.(2009)Krier, Verleysen, Rossi, and Fran{\c
  c}ois]{KrierEtAlESANN2009}
C.~Krier, M.~Verleysen, F.~Rossi, and D.~Fran{\c c}ois.
\newblock Supervised variable clustering for classification of nir spectra.
\newblock In \emph{Proceedings of XVIth European Symposium on Artificial Neural
  Networks (ESANN 2009)}, pages 263--268, Bruges, Belgique, April 2009.

\bibitem[Lechevallier(1976)]{LechevallierContrainte1976}
Y.~Lechevallier.
\newblock Classification automatique optimale sous contrainte d'ordre total.
\newblock Rapport de recherche 200, IRIA, 1976.

\bibitem[Lechevallier(1990)]{LechevallierContrainte1990}
Y.~Lechevallier.
\newblock Recherche d'une partition optimale sous contrainte d'ordre total.
\newblock Rapport de recherche RR-1247, INRIA, June 1990.
\newblock \url{http://www.inria.fr/rrrt/rr-1247.html}.

\bibitem[Lee and Verleysen(2005)]{LeeVerleysenLpWSOM2005}
J.~A. Lee and M.~Verleysen.
\newblock Generalization of the lp norm for time series and its application to
  self-organizing maps.
\newblock In \emph{Proceedings of the 5th Workshop on Self-Organizing Maps
  (WSOM 05)}, pages 733--740, Paris (France), September 2005.

\bibitem[Lin et~al.(2003)Lin, Keogh, Lonardi, and chi
  Chiu]{DBLP:conf/dmkd/LinKLC03}
J.~Lin, E.~J. Keogh, S.~Lonardi, and B.~Y. chi Chiu.
\newblock A symbolic representation of time series, with implications for
  streaming algorithms.
\newblock In M.~J. Zaki and C.~C. Aggarwal, editors, \emph{DMKD}, pages 2--11.
  ACM, 2003.

\bibitem[Picard et~al.(2007)Picard, Robin, Lebarbier, and
  Daudin]{PicardEtAl2007}
F.~Picard, S.~Robin, E.~Lebarbier, and J.-J. Daudin.
\newblock A segmentation/clustering model for the analysis of array cgh data.
\newblock \emph{Biometrics}, 63\penalty0 (3):\penalty0 758--766, 2007.

\bibitem[Ramsay and Silverman(1997)]{RamsaySilverman97}
J.~Ramsay and B.~Silverman.
\newblock \emph{Functional Data Analysis}.
\newblock Springer Series in Statistics. Springer Verlag, June 1997.

\bibitem[Rose(1998)]{rosedeterministicannealing1999}
K.~Rose.
\newblock Deterministic annealing for clustering, compression,classification,
  regression, and related optimization problems.
\newblock \emph{Proceedings of the IEEE}, 86\penalty0 (11):\penalty0
  2210--2239, November 1998.

\bibitem[Rossi and Lechevallier(2008)]{RossiLechevallier2008SFC}
F.~Rossi and Y.~Lechevallier.
\newblock Constrained variable clustering for functional data representation.
\newblock In \emph{Proceedings of the first joint meeting of the
  Soci{\'e}t{\'e} Francophone de Classification and the Classification and Data
  Analysis Group of the Italian Statistical Society (SFC-CLADAG 2008)},
  Caserta, Italy, June 2008.

\bibitem[Rossi et~al.(2004)Rossi, Conan-Guez, and
  El~Golli]{RossiConanGuezElGolliESANN2004SOMFunc}
F.~Rossi, B.~Conan-Guez, and A.~El~Golli.
\newblock Clustering functional data with the {SOM} algorithm.
\newblock In \emph{Proceedings of XIIth European Symposium on Artificial Neural
  Networks (ESANN 2004)}, pages 305--312, Bruges (Belgium), April 2004.

\bibitem[Rossi et~al.(2007)Rossi, Fran{\c c}ois, Wertz, and
  Verleysen]{RossiEtAl06CilsBspline}
F.~Rossi, D.~Fran{\c c}ois, V.~Wertz, and M.~Verleysen.
\newblock Fast selection of spectral variables with b-spline compression.
\newblock \emph{Chemometrics and Intelligent Laboratory Systems}, 86\penalty0
  (2):\penalty0 208--218, April 2007.

\bibitem[Saito and Coifman(1995)]{SaitoCoifman1995LocalBases}
N.~Saito and R.~R. Coifman.
\newblock Local discriminant bases and their applications.
\newblock \emph{Journal of Mathematical Imaging and Vision}, 5\penalty0
  (4):\penalty0 337--358, 1995.

\bibitem[Stone(1961)]{Stone1961}
H.~Stone.
\newblock Approximation of curves by line segments.
\newblock \emph{Mathematics of Computation}, 15\penalty0 (73):\penalty0 40--47,
  1961.

\bibitem[Tikhonov and Arsenin(1979)]{TikhonovArsenin1979}
A.~N. Tikhonov and V.~Y. Arsenin.
\newblock Solutions of ill-posed problems.
\newblock \emph{SIAM Review}, 21\penalty0 (2):\penalty0 266--267, April 1979.

\bibitem[Ultsch and Herrmann(2005)]{DBLP:conf/esann/UltschH05}
A.~Ultsch and L.~Herrmann.
\newblock The architecture of emergent self-organizing maps to reduce
  projection errors.
\newblock In \emph{Proceedings of the 13th European Symposium on Artificial
  Neural Networks (ESANN 2005)}, pages 1--6, Bruges (Belgium), April 2005.

\end{thebibliography}

\end{document}